\documentclass[a4paper]{cas-dc}

\usepackage[numbers]{natbib}
\usepackage{graphicx}
\usepackage{float}
\usepackage{placeins}
\usepackage{caption}

 \newcommand{\add}[1]{{\color{blue}{#1}}}
\renewcommand{\add}[1]{#1}
 
\def\tsc#1{\csdef{#1}{\textsc{\lowercase{#1}}\xspace}}
\tsc{WGM}
\tsc{QE}
\tsc{EP}
\tsc{PMS}
\tsc{BEC}
\tsc{DE}

\begin{document}
\let\WriteBookmarks\relax
\def\floatpagepagefraction{1}
\def\textpagefraction{.001}
\let\printorcid\relax
\shorttitle{IRPol-Fuse: Energy--Structure Coordination for Infrared Polarization Fusion under Low Visibility}
\shortauthors{Huang et al.}

\title [mode = title]{IRPol-Fuse: Energy--Structure Coordination for Infrared Polarization Fusion under Low Visibility}                      

\author[author1]{Zhuangfan~Huang}
\ead{2112455033@stu.fosu.edu.cn}
\fnref{fn1}

\author[author1]{Chusheng~Fang}
\ead{2112555011@stu.fosu.edu.cn}
\fnref{fn1}

\author[author1]{Xiaosong~Li}
\ead{lixiaosong@buaa.edu.cn}
\cormark[1]

\author[author1]{Yang~Liu}
\ead{ly25@fosu.edu.cn}

\author[author2]{Xiaoqi~Cheng}
\ead{chengxiaoqi@fosu.edu.cn}
\cormark[1]

\author[author1]{Haishu~Tan}
\ead{tanhaishu@fosu.edu.cn}

\fntext[fn1]{These authors contributed equally to this work.}
\cortext[cor1]{Corresponding authors.}

\address[author1]{Guangdong-HongKong-Macao Joint Laboratory for Intelligent Micro-Nano Optoelectronic Technology, School of Physics and Optoelectronic Engineering, Foshan University, Foshan 528225, China}
\address[author2]{Guangdong Provincial Key Laboratory of Industrial Intelligent Inspection Technology, Foshan University, Foshan 528225, China.}

\begin{abstract}
Robust perception under low-visibility conditions requires fused imagery that jointly preserves infrared thermal saliency and polarization-derived structural details. However, existing infrared--polarization image fusion (IPIF) methods often overemphasize dominant infrared responses, causing weak yet informative polarization textures in dark regions to be suppressed. To address this issue, we propose IRPol-Fuse, an energy--structure coordinated IPIF framework for challenging low-visibility scenarios. The proposed framework contains three key modules: Polarization Attention Fusion for adaptive infrared--polarization allocation, Infrared Highlight Injector for highlight-guided infrared preservation, and Polarization Texture Injector for polarization texture restoration and fine-detail recovery. We further construct LI-PI, a dedicated infrared--polarization evaluation dataset for low-visibility and visually concealed scenes. Experiments on LI-PI and the public LDDRS dataset demonstrate that IRPol-Fuse achieves favorable performance in thermal target preservation, structural detail recovery, and visual naturalness. Region-aware evaluation and downstream object detection further verify that the proposed energy--structure coordination strategy effectively preserves both infrared target saliency and polarization-derived structural information. Code is available at https://github.com/1hzf/IRPolar-Fuse.
\end{abstract}


\maketitle

\section{Introduction}
Robust perception under low-visibility conditions requires both target saliency and structural completeness.

In this study, low visibility refers to a scene-level condition in which conventional visible observation is degraded by insufficient visible illumination, local glare, occlusion, or visual concealment. This term does not imply that long-wave infrared imaging depends on ambient visible illumination, since LWIR sensors primarily capture thermally emitted radiation from objects rather than reflected visible light~\cite{54}.

However, single-modality imaging is often insufficient in such scenarios. Visible imaging is easily degraded by weak illumination and scattering, leading to texture loss and boundary blur~\cite{1}, whereas infrared imaging, although effective in highlighting thermally salient targets, remains limited in representing fine structures and high-frequency details~\cite{2}. Moreover, when the target and background approach thermal equilibrium, the discriminability of infrared images is further reduced~\cite{3}.

\begin{figure}
\centering
\includegraphics[width=1\linewidth]{ 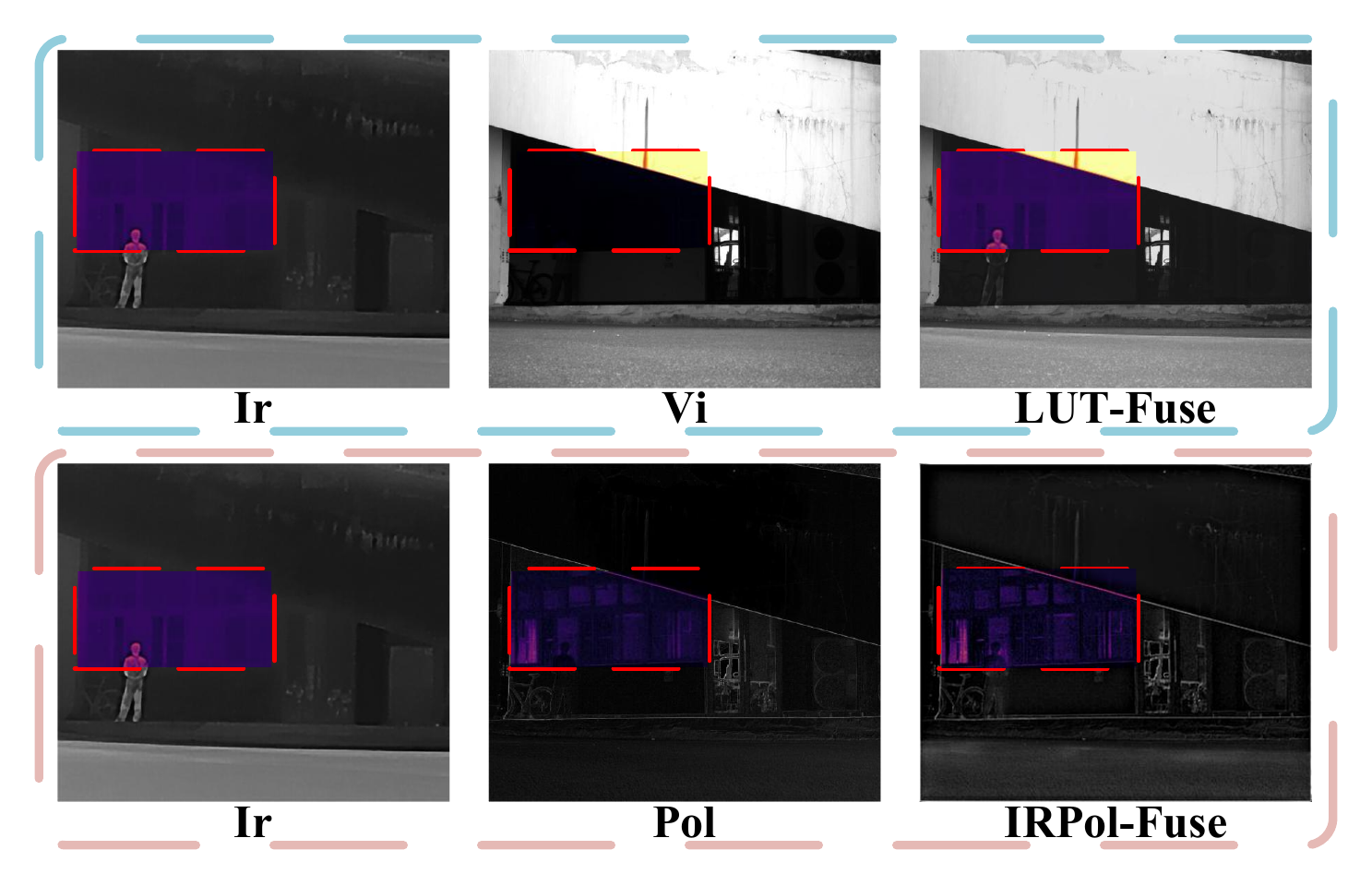}
\vspace{-1em}
\caption{Comparison of IVIF and IPIF.}
\vspace{-1em}
\label{fig1}
\end{figure}

Recent infrared-only methods have achieved promising performance in task-specific perception, such as instance segmentation, non-contact temperature measurement, and small-target detection~\cite{45,46,47}. However, these methods still mainly rely on thermal contrast and provide limited surface-texture information. Low-light enhancement methods can improve degraded visible images through state-space modeling, Retinex decomposition, wavelet representation, and luminance-guided Transformers~\cite{48,49}, but they cannot introduce missing thermal cues when targets are visually concealed. Infrared--visible fusion methods further combine thermal saliency with visible textures~\cite{50,51,52}; nevertheless, their structural contribution remains dependent on the quality of the visible input and may become unreliable under severe visibility degradation.

These limitations motivate the introduction of polarization imaging. Polarization imaging provides complementary cues that are more sensitive to surface reflection properties, material characteristics, and fine geometric structures than conventional intensity imaging~\cite{4}, making it a promising complement to infrared imaging for low-visibility perception.

As shown in Fig.~\ref{fig1}, the visible image suffers from severe texture degradation in the window region under weak illumination, and the corresponding infrared-visible fusion result still fails to recover structural details in dark areas. By contrast, the polarization image preserves clearer window-frame boundaries and internal texture patterns, which are more effectively retained in the corresponding infrared--polarization fusion result. Moreover, polarization information is complementary to conventional intensity information in its representation mechanism~\cite{5}. These observations suggest that, compared with infrared-visible image fusion (IVIF), IPIF is better suited for jointly preserving thermal saliency and structural completeness in challenging low-visibility environments.

\begin{figure}
    \includegraphics[width=1\linewidth]{ 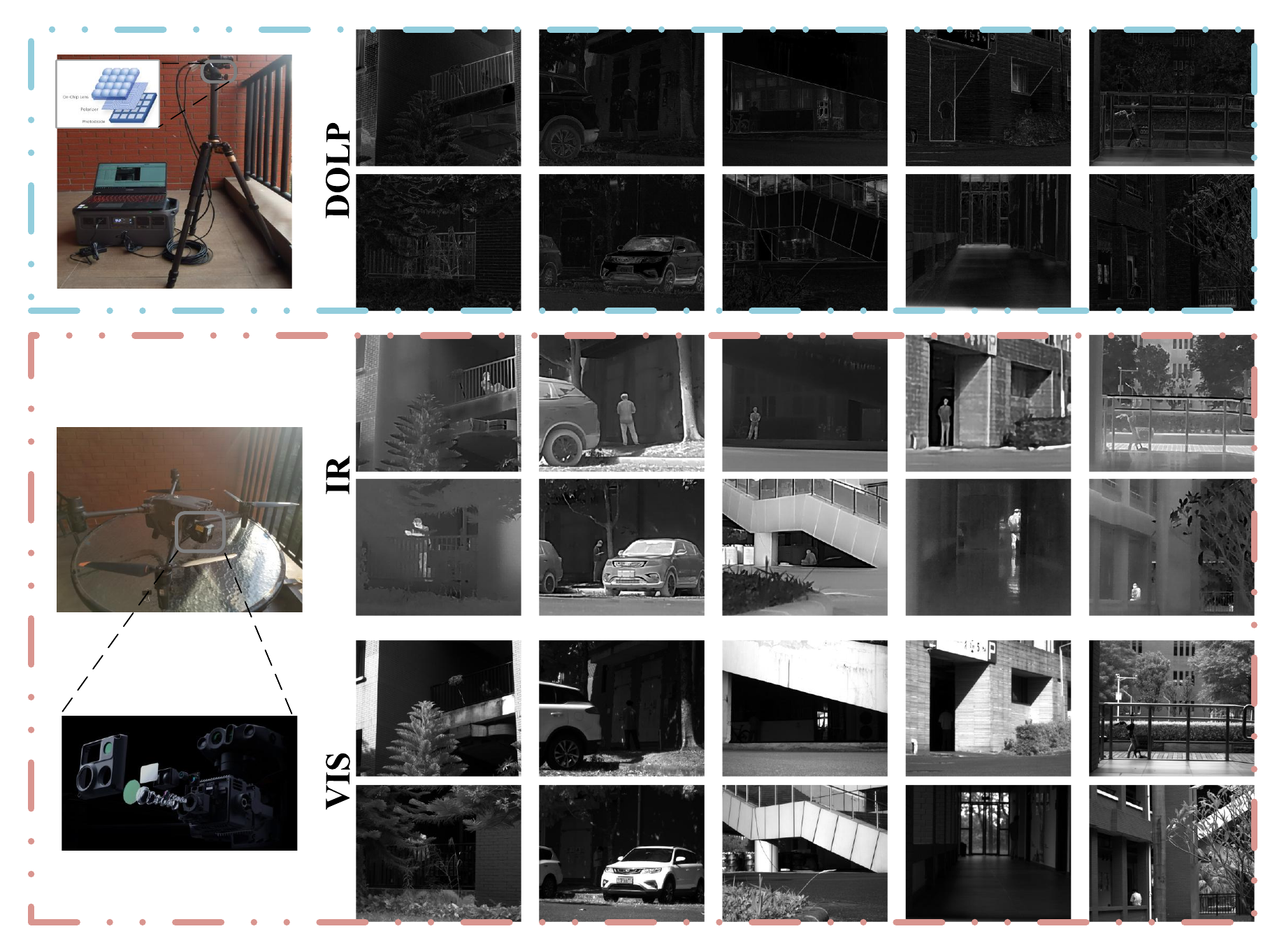}
    \vspace{-2em}
    \caption{Illustration of the LI-PI dataset. }
    \vspace{-2em}
    \label{fig2}
\end{figure}

Deep learning-based IPIF methods have improved cross-modal representation and global contextual modeling~\cite{6,8,9}. For example, TIPFNet introduces the Swin Transformer for $DoLP$ and $S_{0}$ fusion with gradient-residual fusion and self-supervised constraints~\cite{7}. However, existing methods may still overemphasize dominant infrared responses and suppress weak polarization-derived textures in challenging low-visibility and visually concealed scenes. To address this issue, we propose IRPol-Fuse, an energy--structure coordinated IPIF framework for infrared---polarization fusion under low-visibility conditions. The main contributions are summarized as follows:

\relpenalty=10000
\binoppenalty=10000
\sloppy
\begin{enumerate}
\itemsep = 0pt

\item We identify a key limitation of existing IPIF methods in challenging low-visibility scenarios: they often over-favor dominant infrared responses while insufficiently preserving weak polarization-derived structures in degraded or visually concealed regions. To address this issue, we propose IRPol-Fuse, an infrared--polarization fusion framework tailored for low-visibility and visually concealed environments.

\item We introduce an energy--structure coordinated fusion design, including Polarization Attention Fusion (PAF), Infrared Highlight Injector (IHJ), and Polarization Texture Injector (PTJ), to balance infrared target preservation and polarization texture recovery. A selective state-space interaction mechanism is further used to enhance long-range cross-modal modeling.

\item LI-PI is constructed as a dedicated infrared--polarization evaluation dataset for low-visibility and visually concealed scenarios, as illustrated in Fig.~\ref{fig2}. With 110 strictly aligned samples, it provides a targeted evaluation setting for studying the coordinated preservation of infrared saliency and polarization structure under challenging scene conditions.
\end{enumerate}  

The remainder of this paper is organized as follows. Section 2 reviews related studies on IPIF. Section 3 details the proposed IRPol-Fuse framework and the LI-PI dataset. Section 4 presents experimental settings, comparative results, and ablation analyses. Finally, Section 5 concludes the paper.

\section{Related work}
\subsection{Stokes-Based Polarization Representation}
Polarization representation provides the physical basis for IPIF. Among existing formulations, the Stokes-vector representation is widely adopted because it offers a unified description of polarization states and can be directly derived from multi-orientation intensity measurements. The corresponding Stokes formulation and $DoLP$ definition are given in Eqs.~\eqref{eq1}--\eqref{eq3}.

\begin{equation}
\label{eq1}
\hspace{+8mm}
S  =  \left[\begin{array}{l}
S_{0} \\
S_{1} \\
S_{2} \\
S_{3}
\end{array}\right]  
= \left[\begin{array}{c}
I_{H}+I_{V} \\
I_{H}-I_{V} \\
I_{45}-I_{135} \\
I_{R}-I_{L}
\end{array}\right]
\end{equation}

\begin{equation}
\label{eq2}
\hspace{+9mm}
q = \frac{S_{1}}{S_{0}}, \qquad
u = \frac{S_{2}}{S_{0}}
\end{equation}

\begin{equation}
\label{eq3}
\hspace{+8mm}
DoLP = \frac{\sqrt{S_{1}^{2}+S_{2}^{2}}}{S_{0}}
      = \sqrt{q^{2}+u^{2}}
\end{equation}

For the low-visibility IPIF task, infrared intensity and polarization representation provide complementary cues. Infrared images mainly emphasize thermal saliency, whereas DoLP is more sensitive to surface textures, local structures, and material-related boundaries. Therefore, DoLP offers structural information that is particularly valuable in dark and low-contrast regions, which forms the representation basis of the energy–structure coordination problem addressed in this work.

\subsection{Overview of the Polarized Image Fusion Methods}

Existing IPIF methods can generally be categorized into conventional methods and deep learning-based methods. Conventional methods mainly include multiscale analysis~\cite{10}, sparse representation~\cite{11}, and saliency-driven strategies~\cite{13}. These methods typically project heterogeneous modalities into a unified feature space through pyramid decomposition, transform-domain representation, or hand-crafted fusion rules, followed by image reconstruction. Although they provide relatively strong physical interpretability and achieved promising results in early studies, their fixed fusion rules often struggle to simultaneously preserve thermally salient targets and restore high-frequency structural details, especially when infrared and polarization modalities differ substantially in statistical distribution and physical characteristics~\cite{12}.

Due to their strong feature extraction and cross-modal representation capabilities, deep learning-based methods have been widely adopted in polarization fusion, but their local receptive fields limit long-range dependency modeling and global contextual understanding~\cite{4,14}. Generative adversarial networks can improve the visual realism of fused results through adversarial constraints, yet their practical deployment remains limited by unstable optimization and mode collapse~\cite{18,21}. By contrast, Transformer-based methods capture cross-modal global dependencies through self-attention and therefore show stronger potential for structural detail preservation and contextual modeling~\cite{22,23}. Nevertheless, under low-visibility conditions, effectively preserving infrared thermal saliency while recovering weak polarization-derived structural details remains a key challenge in this field.

\begin{figure*}
    \includegraphics[width=1\linewidth]{ 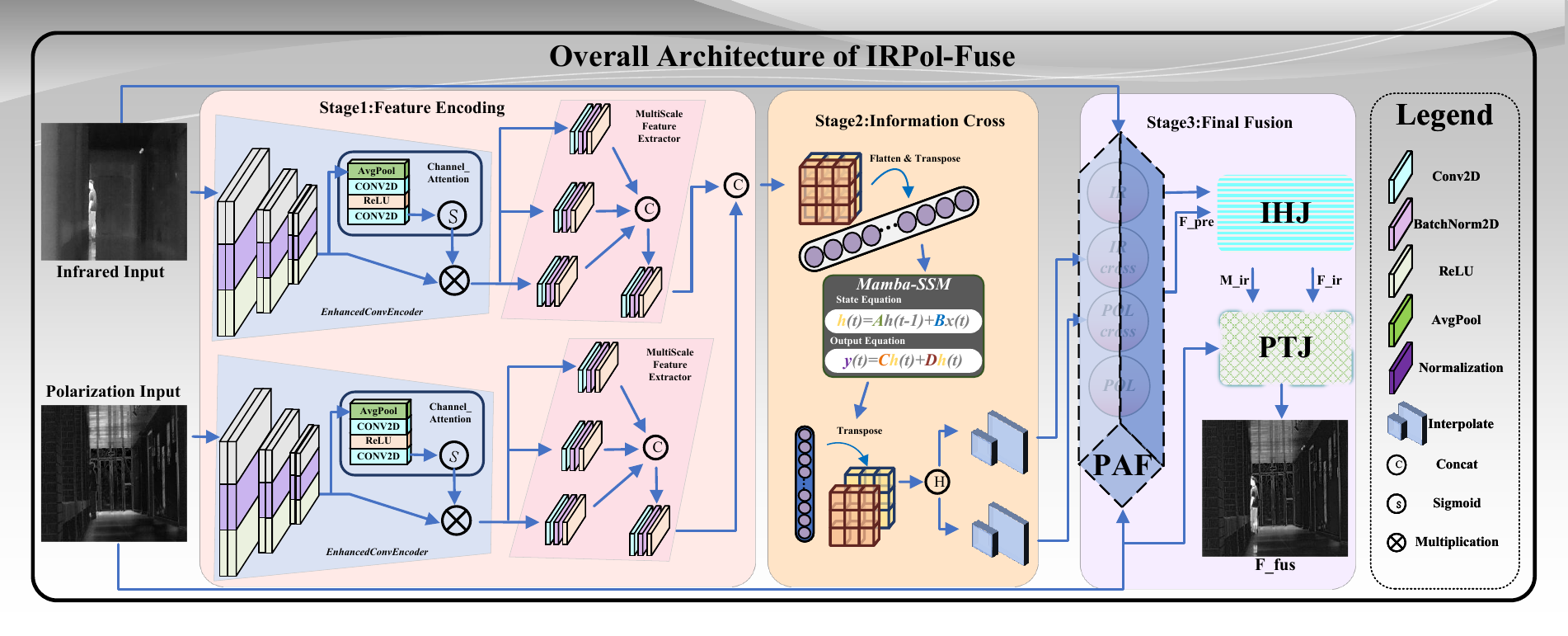}
    \vspace{-2em}
    \caption{Overall architecture of IRPol-Fuse.}
    \vspace{-2em}
    \label{overall}
\end{figure*}
\begin{figure*}
    \includegraphics[width=1\linewidth]{ 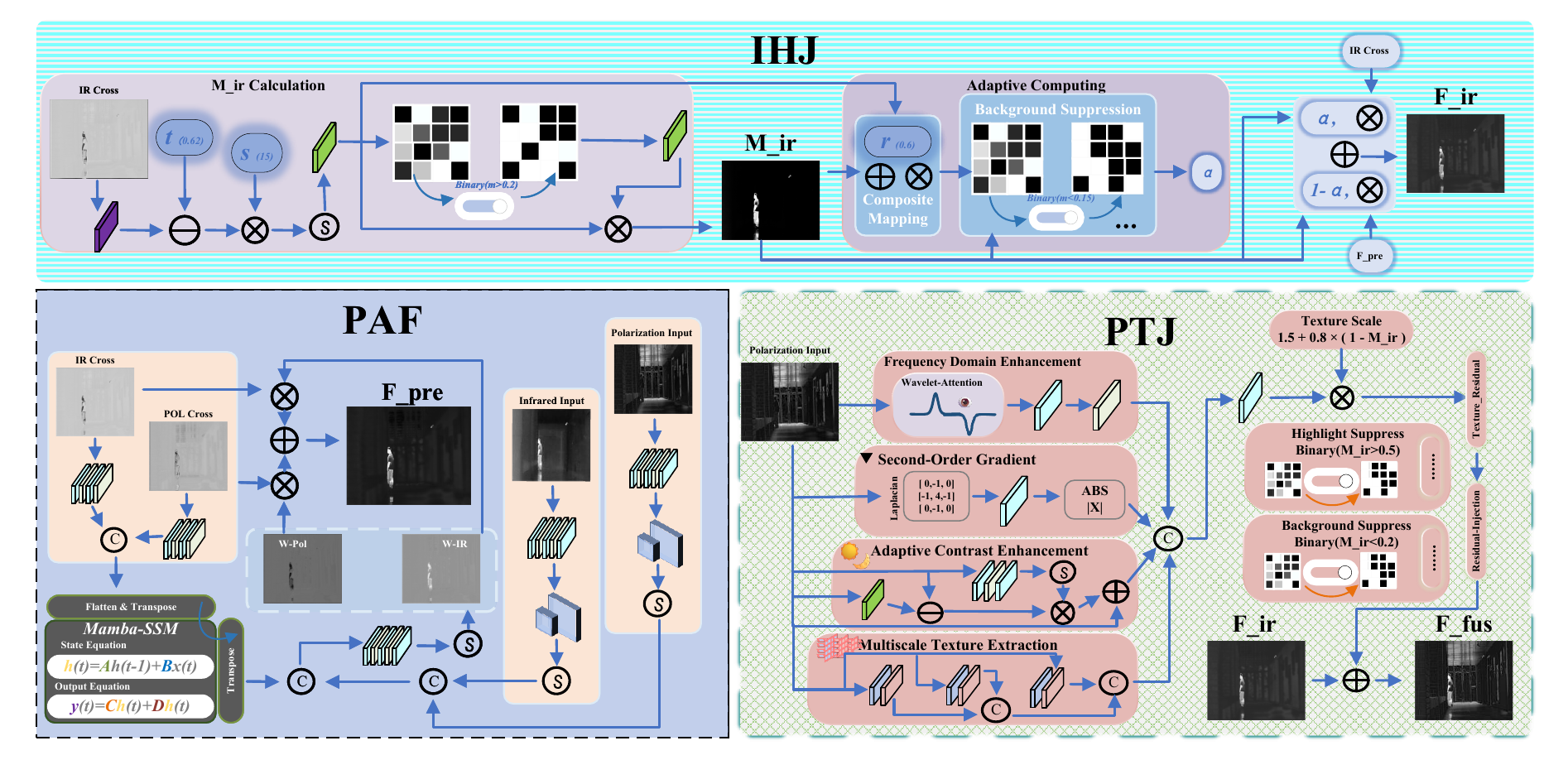}
    \vspace{-2em}
    \caption{Detailed structures of the proposed PAF, IHJ, and  PTJ modules.}
    \vspace{-2em}
    \label{module}
\end{figure*}

\subsection{Development Trends of Infrared Polarization}
Infrared intensity images and polarization information characterize complementary physical attributes of a target: the former mainly reflects thermal radiative energy and target saliency, whereas the latter is more sensitive to surface microstructures, roughness, and geometric morphology. Owing to this complementarity, infrared--polarization fusion has shown promising potential in applications such as low-visibility perception~\cite{26}, target detection~\cite{1}, material recognition~\cite{5}, and scene analysis under complex backgrounds~\cite{28}. Representative studies such as TIPFNet introduce Transformer-based global modeling into the fusion of DoLP and $S_0$, improving the joint representation of salient targets and background details through long-range dependency modeling and gradient-residual fusion~\cite{7}. Subsequent efforts have further incorporated polarization priors, low-rank representation, and guided filtering to better coordinate infrared energy preservation and polarization texture restoration~\cite{12}. These developments indicate that infrared--polarization fusion is gradually evolving from generic data-driven fusion toward more physically informed and context-aware frameworks.

However, existing IPIF methods remain insufficient for more challenging
scenarios involving degraded visible observation, local glare interference,
and visually concealed targets. Most current frameworks still tend to favor dominant infrared responses, causing weak yet informative polarization textures in dark regions to be suppressed during fusion, particularly those related to boundaries, local textures, and material-specific structures. Consequently, although thermal targets may remain salient, fine structural recovery and boundary continuity are often inadequate. To address these limitations, we develop IRPol-Fuse as an IPIF framework for challenging low-visibility scenarios, which coordinates cross-modal collaborative modeling, thermal target preservation, and
polarization texture enhancement in a unified manner.

\section{Proposed Methods}
\subsection{Overall Architecture}

IRPol-Fuse is an energy--structure coordinated infrared--polarization image fusion framework designed for challenging low-visibility scenarios. As illustrated in Fig.~\ref{overall}, the overall architecture follows a three-stage pipeline, including modality-specific feature encoding, cross-modal Mamba interaction, and sequential fusion through PAF, IHJ, and  PTJ modules.

Given an infrared image $I_{\mathrm{ir}}$ and a polarization image $I_{\mathrm{pol}}$, IRPol-Fuse first extracts modality-specific shallow and multi-scale features from the two input branches. The extracted features are then fed into a cross-modal Mamba interaction module to model long-range dependencies and complementary information between infrared thermal responses and polarization structural cues. After cross-modal interaction, PAF generates a preliminary fused result $F_{\mathrm{pre}}$ through adaptive infrared--polarization allocation. IHJ further produces an infrared--enhanced fused result $F_{\mathrm{ir}}$ by reinjecting thermally salient infrared responses. Finally, PTJ restores polarization-derived texture details and generates the final fused output $F_{\mathrm{fus}}$.

The detailed structures of PAF, IHJ, and PTJ are provided in Fig.~\ref{module} and described separately in the following subsections.

\subsection{Polarization Attention Fusion (PAF)}

PAF is designed to generate a preliminary fused representation by adaptively integrating infrared thermal responses and polarization structural information after cross-modal interaction. As illustrated in Fig.~\ref{module}, the cross-modal features from the infrared and polarization branches are denoted as $F_{\mathrm{IR}}^{c}$ and $F_{\mathrm{POL}}^{c}$, respectively. They are first projected into two modality-specific response maps, denoted as $W_{\mathrm{IR}}$ and $W_{\mathrm{POL}}$. These response maps are produced by lightweight prediction heads consisting of two convolutional layers followed by a Sigmoid activation, enabling spatially adaptive weighting within the range of $[0,1]$.

Unlike direct feature addition, PAF performs region-adaptive energy--structure allocation by dynamically regulating the contribution of each modality according to the local feature responses. Specifically, regions with strong thermal responses receive higher infrared weights, whereas texture-rich regions preserve stronger polarization contributions. The preliminary fused representation is therefore formulated as

\begin{equation}
F_{\mathrm{pre}}
=
W_{\mathrm{IR}}\odot F_{\mathrm{IR}}^{c}
+
W_{\mathrm{POL}}\odot F_{\mathrm{POL}}^{c},
\label{eq:paf}
\end{equation}
where $\odot$ denotes element-wise multiplication. $F_{\mathrm{IR}}^{c}$ and $F_{\mathrm{POL}}^{c}$ denote the cross-modal infrared and polarization features, respectively. The weighting maps are generated from the corresponding modality features rather than predefined coefficients, allowing the fusion process to adapt automatically to different scene characteristics.

Since PAF only performs preliminary feature aggregation, thermal responses in highly radiant regions may still be weakened by cross-modal mixing. Therefore, the generated preliminary fusion result $F_{\mathrm{pre}}$ is subsequently refined by the proposed IHJ, which explicitly preserves thermally salient infrared information before polarization texture restoration.

\subsection{ Infrared Highlight Injector (IHJ)}

Although PAF provides an effective preliminary fusion, thermally salient infrared responses may still be weakened during cross-modal feature aggregation. To explicitly preserve high-radiance targets while suppressing unnecessary thermal reinjection into background regions, the proposed IHJ performs adaptive infrared compensation according to an infrared highlight mask.

As illustrated in Fig.~\ref{module}, the infrared image is first normalized into a relative intensity range. A lightweight soft-threshold function is then employed to generate an initial infrared highlight response,

\begin{equation}
M_{o}=\sigma\left(s\left(I_{\mathrm{ir}}^{n}-t\right)\right),
\end{equation}
where $I_{\mathrm{ir}}^{n}$ denotes the normalized infrared intensity map, and
$t$ and $s$ are fixed IHJ hyperparameters controlling the activation boundary
and transition sharpness, respectively. After lightweight smoothing and
small-kernel mask refinement, $M_{o}$ is converted into the final infrared
highlight mask $M_{\mathrm{ir}}$.

The Composite Mapping operation is then explicitly defined as
\begin{equation}
\alpha_{0}
=
r\left(
M_{\mathrm{ir}}
+
0.4M_{\mathrm{ir}}^{2}
\right),
\end{equation}
where $r$ is the fixed base injection ratio. The linear mask term provides
smooth infrared compensation in transition regions, while the quadratic term
increases the response for thermally salient regions. To suppress unnecessary
infrared reinjection in weak-response background regions, a binary background
mask is further applied in a pixel-wise manner. Specifically, when
$M_{\mathrm{ir}}<0.15$, the adaptive coefficient is reduced to
$0.2\alpha_{0}$; otherwise, $\alpha_{0}$ is retained. The resulting
background-suppressed coefficient is denoted as $\alpha$.

The infrared--enhanced intermediate fusion result is formulated as
\begin{equation}
F_{\mathrm{ir}}
=
\left(
1-\alpha M_{\mathrm{ir}}
\right)
\odot F_{\mathrm{pre}}
+
\alpha M_{\mathrm{ir}}
\odot I_{\mathrm{ir}}.
\end{equation}
This formulation shows that the preliminary fused result $F_{\mathrm{pre}}$ is
selectively corrected by the source infrared image $I_{\mathrm{ir}}$ under the
joint guidance of the highlight mask $M_{\mathrm{ir}}$ and the adaptive injection
coefficient $\alpha$. In this way, IHJ preserves thermally salient infrared
responses while avoiding unnecessary infrared reinjection into background regions.

\subsection{Polarization Texture Injector (PTJ)}

Although IHJ effectively preserves thermally salient infrared responses, the adaptive infrared reinjection inevitably suppresses part of the fine structural information originating from the polarization modality. Therefore, the proposed PTJ is introduced to restore polarization-derived texture details while maintaining thermal saliency. As illustrated in Fig.~\ref{module}, PTJ consists of four complementary branches, namely Frequency Domain Enhancement, Second-Order Gradient, Adaptive Contrast Enhancement, and Multiscale Texture Extraction. These branches jointly extract frequency-domain, structural, contrast-aware, and multi-scale texture information from the polarization image to construct a comprehensive texture residual.

Specifically, the Frequency Domain Enhancement branch performs Haar wavelet decomposition on the polarization image to obtain four frequency subbands, including LL, LH, HL, and HH. The LL subband preserves coarse structural and contextual information, whereas the LH, HL, and HH subbands contain horizontal, vertical, and diagonal high-frequency responses associated with edges and fine textures. Channel attention is applied to all four subbands to recalibrate their frequency responses, while spatial attention is further applied to the high-frequency subbands to emphasize texture-rich regions. The enhanced frequency representation is reconstructed through inverse Haar transform and can be compactly expressed as

\begin{equation}
F_{\mathrm{freq}}
=
\mathcal{D}_{\mathrm{Haar}}^{-1}
\!\left(
\mathcal{A}
\!\left(
\mathcal{D}_{\mathrm{Haar}}
(I_{\mathrm{pol}})
\right)
\right),
\label{eq:ptj1}
\end{equation}
where $\mathcal{D}_{\mathrm{Haar}}(\cdot)$ and $\mathcal{D}_{\mathrm{Haar}}^{-1}(\cdot)$ denote the Haar wavelet decomposition and inverse reconstruction, respectively, and $\mathcal{A}(\cdot)$ represents the attention-guided frequency enhancement operation.

Besides frequency-domain enhancement, PTJ further extracts complementary structural information using the second-order gradient, adaptive contrast enhancement, and multi-scale texture extraction branches. The second-order gradient branch emphasizes structural discontinuities through Laplacian responses, the adaptive contrast branch enhances weak local polarization variations, and the multi-scale branch captures texture information under different receptive fields. The outputs of the four branches are aggregated to generate the polarization texture residual,

\begin{equation}
R_{\mathrm{tex}}
=
\phi
\left(
F_{\mathrm{freq}},
F_{\mathrm{grad}},
F_{\mathrm{acc}},
F_{\mathrm{ms}}
\right),
\label{eq:ptj2}
\end{equation}
where $F_{\mathrm{freq}}$, $F_{\mathrm{grad}}$, $F_{\mathrm{acc}}$, and $F_{\mathrm{ms}}$ denote the frequency-domain enhanced feature, second-order gradient feature, adaptive contrast-enhanced feature, and multi-scale texture feature, respectively, and $\phi(\cdot)$ represents feature concatenation followed by convolutional aggregation.

To maintain consistency with the adaptive infrared compensation introduced by IHJ, the generated texture residual is further modulated according to the infrared highlight mask before residual injection. Specifically, texture enhancement is suppressed in thermally dominant regions while relatively stronger structural compensation is preserved in weak-response background regions. This mask-guided modulation can be summarized as

\begin{equation}
\tilde{R}_{\mathrm{tex}}
=
S
\left(
R_{\mathrm{tex}},
M_{\mathrm{ir}}
\right),
\qquad
F_{\mathrm{fus}}
=
F_{\mathrm{ir}}
+
\tilde{R}_{\mathrm{tex}},
\label{eq:ptj3}
\end{equation}
where $S(\cdot)$ denotes the mask-guided texture modulation, including
texture scaling, highlight suppression, and background enhancement. Specifically,
texture responses are suppressed in thermally dominant regions, while
polarization-derived structural compensation is enhanced in low-infrared-response
regions. By jointly integrating frequency-domain enhancement, structural gradient
extraction, adaptive contrast enhancement, and multi-scale texture aggregation,
PTJ effectively restores polarization-derived structural details without
introducing excessive texture responses into thermally salient infrared regions.

\subsection{Loss Function Design}

To guide energy--structure coordinated fusion, IRPol-Fuse is optimized by a multi-objective loss function consisting of global fusion consistency, infrared target preservation, and polarization texture restoration terms:

\begin{equation}
\mathcal{L}_{\mathrm{total}}
=
\lambda_{\mathrm{fusion}}\mathcal{L}_{\mathrm{fusion}}
+
\lambda_{\mathrm{ir}}\mathcal{L}_{\mathrm{ir}}
+
\lambda_{\mathrm{pol}}\mathcal{L}_{\mathrm{pol}},
\label{eq:loss_total}
\end{equation}
where $\lambda_{\mathrm{fusion}}$, $\lambda_{\mathrm{ir}}$, and $\lambda_{\mathrm{pol}}$ are weighting coefficients for global fusion consistency, infrared target preservation, and polarization texture restoration, respectively. In the final configuration, we set $\lambda_{\mathrm{fusion}}=1.0$, $\lambda_{\mathrm{ir}}=2.0$, and $\lambda_{\mathrm{pol}}=2.5$.

The fusion consistency loss constrains the fused result from both intensity and structural perspectives:
\begin{equation}
\begin{gathered}
\mathcal{L}_{\mathrm{fusion}}
=
\mathcal{L}_{\mathrm{int}}
+
\mathcal{L}_{\mathrm{grad}},
\\[2mm]
\mathcal{L}_{\mathrm{int}}
=
\mathrm{Nor}
\left(
F_{\mathrm{fus}}
-
I_{\mathrm{ir}}
\right)
+
\mathrm{Nor}
\left(
F_{\mathrm{fus}}
-
I_{\mathrm{pol}}
\right),
\\[2mm]
\mathcal{L}_{\mathrm{grad}}
=
\mathrm{Nor}
\left(
\nabla F_{\mathrm{fus}}
-
\max
(
\nabla I_{\mathrm{ir}},
\nabla I_{\mathrm{pol}}
)
\right).
\end{gathered}
\label{eq:loss_fusion}
\end{equation}

where $\mathrm{Nor}(\cdot)$ denotes normalization and $\nabla(\cdot)$ denotes the first-order gradient operator.

The infrared--preserving loss is introduced to preserve thermally salient targets during fusion. A highlight-adaptive mask $M_{h}$ is generated from high-radiance regions in the infrared image, and the loss is defined as
\begin{equation}
\begin{aligned}
\mathcal{L}_{\mathrm{ir}}
=
&
\,3\,
\mathrm{Nor}
\left(
M_h
\odot
(F_{\mathrm{fus}}-I_{\mathrm{ir}})
\right)
\\
&
+
1.5\,
\mathrm{Nor}
\left(
(1-M_h)
\odot
(F_{\mathrm{fus}}-I_{\mathrm{pol}})
\right).
\end{aligned}
\label{eq:loss_ir}
\end{equation}

The first term preserves thermal energy in highlight regions, whereas the second term introduces polarization guidance in non-highlight areas. Thus, $\mathcal{L}_{\mathrm{ir}}$ provides a region-adaptive thermal constraint.

The polarization-preserving loss maintains high-frequency textures and weak structural information from the polarization modality. A dark-region adaptive mask $M_{bk}$ is obtained from low-intensity regions of the polarization image, and the polarization loss is formulated as

\begin{equation}
\mathcal{L}_{\mathrm{pol}}
=
\mathrm{Nor}
\left(
\nabla^{2}F_{\mathrm{fus}}
-
\nabla^{2}I_{\mathrm{pol}}
\right)
+
\mathrm{Nor}
\left(
M_{bk}\odot(F_{\mathrm{fus}}-I_{\mathrm{pol}})
\right),
\label{eq:loss_pol}
\end{equation}
where $\nabla^{2}(\cdot)$ denotes the Laplacian operator. The first term preserves second-order structural responses, while the second term strengthens polarization feature preservation in weak-texture regions.

Overall, $\mathcal{L}_{\mathrm{fusion}}$ enforces global consistency, $\mathcal{L}_{\mathrm{ir}}$ preserves thermally salient targets, and $\mathcal{L}_{\mathrm{pol}}$ restores polarization-derived textures and fine structures. Their joint optimization encourages fused results with improved target saliency, structural fidelity, and visual naturalness.

\subsection{LI-PI Dataset}

Publicly available infrared--polarization datasets still suffer from limited scene diversity and illumination complexity. For example, the LDDRS dataset mainly focuses on autonomous driving and road detection scenarios and can reasonably reflect infrared--polarization imaging characteristics in conventional outdoor environments~\cite{33}. However, its samples are relatively homogeneous and are mostly collected under stable illumination conditions, making it insufficient for more challenging scenarios involving degraded visible observation, local glare interference, occlusion, and visually concealed targets.

To address this limitation, LI-PI is constructed as a dedicated infrared--polarization evaluation dataset for challenging low-visibility conditions. It covers representative scenes such as parking garages, balconies, forests, and corridors, and includes degraded visible observation, local glare interference, and visually concealed targets. In the visually concealed setting, the human target is difficult to observe in visible imagery, while remaining thermally salient in infrared images and preserving certain structural contours in polarization images. This design makes LI-PI a targeted evaluation dataset for studying the coordinated preservation of infrared saliency and polarization structure in realistic low-visibility environments.

LI-PI contains 110 strictly aligned infrared--polarization image pairs. 
We randomly divide it into 90 pairs for training, 10 pairs for validation, 
and 10 pairs for independent testing. The validation set is used for model 
selection and hyperparameter analysis, while the independent test set is used 
only for final evaluation. No image pair is shared across the three subsets.
Although LI-PI provides a targeted evaluation setting for low-visibility 
infrared--polarization fusion, its scale is still limited. Constructing a larger 
and more diverse infrared--polarization benchmark remains an important direction 
for future work.

\section{Experiments}
\subsection{Experimental Settings}

To ensure a standardized and reproducible evaluation, IRPol-Fuse is implemented in PyTorch and trained on a single NVIDIA GeForce RTX 3090 GPU for both training and inference. The network parameters are optimized using Adam with an initial learning rate of $1\times10^{-4}$.

For LI-PI, we use the same 90/10/10 split described in Section 3.6, 
where 90 image pairs are used for training, 10 for validation, and 10 for 
independent testing. All quantitative comparisons, ablation studies, 
region-aware evaluations, and qualitative visualizations on LI-PI are conducted 
on the independent test set. All competing methods are evaluated using the same 
test images and the same preprocessing protocol.

\add{
For comparison, nine recent and publicly reproducible methods are selected as baselines, including representative infrared--polarization fusion approaches and multimodal image fusion frameworks with dedicated polarization fusion settings, namely TIPFNet~\cite{7}, CPIFuse~\cite{34}, DT-F~\cite{23}, PIPFNet~\cite{12}, LFDT~\cite{37}, FusionMamba~\cite{38}, CDDFuse~\cite{39}, LUT-Fuse~\cite{40}, and SeAFusion~\cite{41}.} In particular, CPIFuse, DT-F, PIPFNet, and LFDT are newly included to reflect recent developments in polarization image fusion. For a fair comparison, all competing methods are reproduced and evaluated using the same source images, preprocessing settings, and quantitative evaluation protocol. The same comparison set is consistently adopted in all quantitative tables and corresponding qualitative figures.

For quantitative evaluation, we adopt complementary global fusion metrics and region-aware metrics. The global metrics include $Q_{P}$, $Q_{S}$, $Q_{CB}$, $Q_{CV}$, $Q_{AB/F}$, and MS-SSIM, which evaluate source-information preservation, structural consistency, perceptual quality, fusion distortion, edge transfer, and multi-scale structural similarity, respectively. Together, these metrics provide an overall assessment of image-level fusion quality.

To further validate the proposed energy--structure coordination strategy,
four region-aware metrics are introduced. The infrared target ROI is automatically
generated from the source infrared image $I_{\mathrm{ir}}$, rather than from the
fused results. Specifically, $I_{\mathrm{ir}}$ is normalized to $[0,1]$, and
pixels satisfying $I_{\mathrm{ir}}^{n}>0.62$ are selected as the infrared ROI.
If the selected region contains fewer than 100 pixels, the highest-response
infrared pixels are used as a fallback ROI to avoid unstable statistics. The
complementary non-highlight region is used for polarization-texture evaluation.
All compared methods use the same ROI masks, ensuring that the region-aware
evaluation is independent of the fusion results.

IR ROI Corr and IR CNR are computed within the infrared ROI to evaluate infrared
target preservation, while Pol Grad Corr and Pol Grad MAE are computed within
the complementary non-highlight region to assess polarization texture recovery.

\subsection{Analysis of Experimental Results}

\begin{figure*}[!t]
    \centering
    \centering
    \includegraphics[width=1\linewidth]{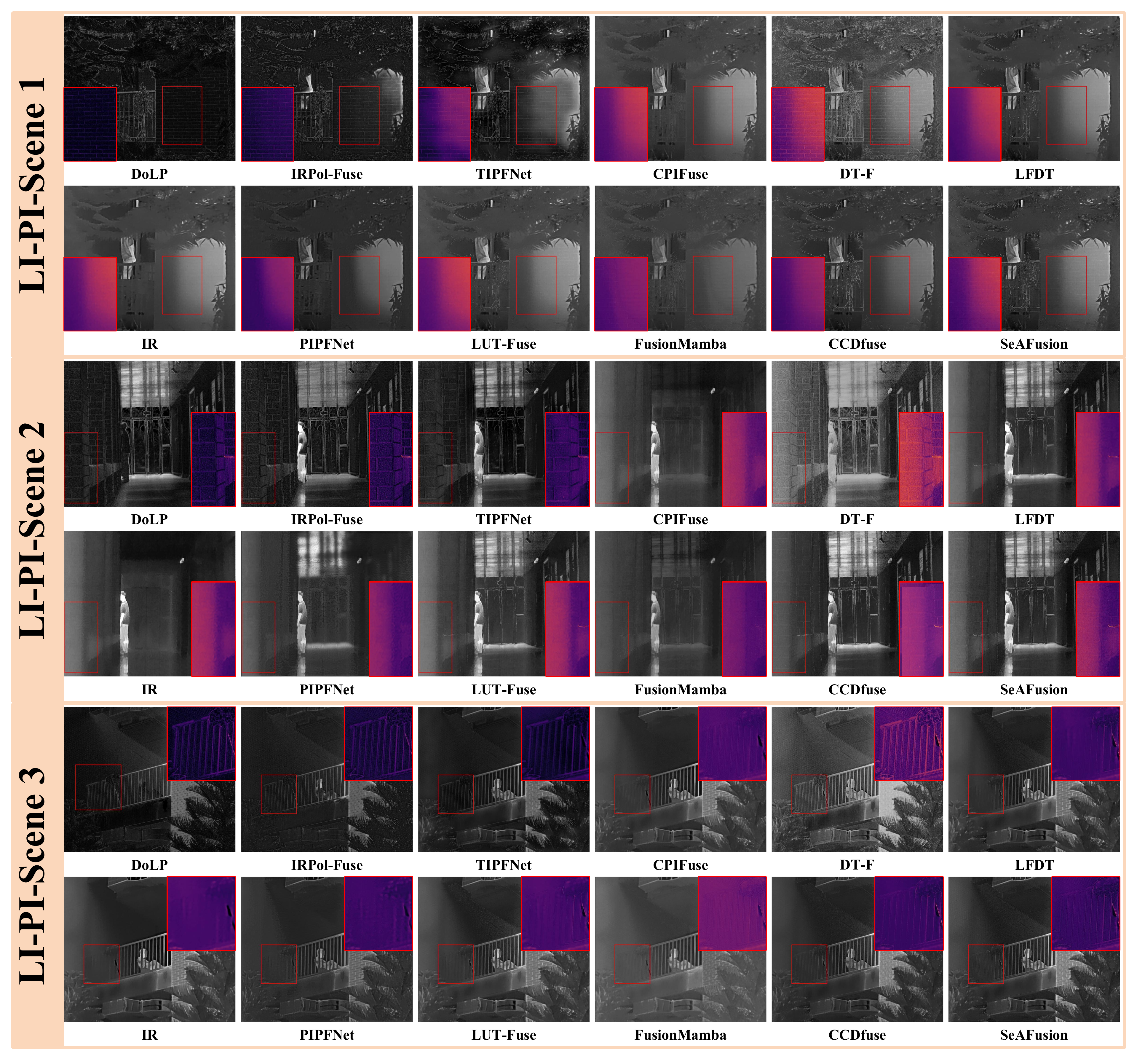}
    \vspace{-2em}
    \caption{Comparison of fusion results in three representative low-visibility scenes from the LI-PI dataset.}
    \vspace{-2em}
    \label{fig:lipi_results}
\end{figure*}

To evaluate the fusion performance of IRPol-Fuse under challenging low-visibility conditions, we conduct qualitative and quantitative comparisons on the LI-PI dataset. As shown in Fig.~\ref{fig:lipi_results}, the selected scenes contain visually concealed targets, local glare interference, weak background textures, and complex structural details, which are suitable for examining infrared target preservation and polarization texture recovery.

In LI-PI Scene 1, the background contains rich polarization textures together with redundant infrared responses. Some comparison methods, such as TIPFNet, CPIFuse, DT-F, and LFDT, preserve part of the salient thermal information, but tend to weaken fine wall textures or introduce excessive infrared dominance in non-target regions. LUT-Fuse, FusionMamba, CDDFuse, and SeAFusion retain certain global structures but show insufficient recovery of local polarization details. By contrast, IRPol-Fuse better preserves the human thermal contour while maintaining clearer wall textures, indicating a more balanced energy--structure coordination.

In LI-PI Scene 2, the target is visually concealed and the background contains weak structural cues. Several competing methods either over-enhance infrared responses or suppress polarization-derived background details. IRPol-Fuse maintains clearer target boundaries and stronger wall-structure continuity, showing that the proposed IHJ and PTJ modules can jointly preserve thermal saliency and recover polarization textures.

In LI-PI Scene 3, strong highlight interference and complex railing structures make it difficult to simultaneously preserve infrared energy and fine polarization details. Methods such as CPIFuse, DT-F, and LFDT tend to emphasize infrared responses, whereas some other methods fail to maintain sufficient railing structures. IRPol-Fuse suppresses highlight interference more effectively and restores clearer railing contours and background structures, demonstrating the benefit of the proposed texture residual modulation.

The quantitative results in Table~\ref{table:lipi_global} are consistent with the qualitative observations. IRPol-Fuse achieves the best performance on all six global metrics, including $Q_{P}$, $Q_{S}$, $Q_{CB}$, $Q_{CV}$, $Q_{AB/F}$, and MS-SSIM. These results indicate that the proposed method provides better source-information preservation, structural consistency, perceptual quality, edge transfer, and multi-scale structural similarity on the LI-PI dataset.

\begin{table}[!t]
\centering
\small
\setlength{\tabcolsep}{3pt}
{\rmfamily
\caption{Quantitative comparison results on the LI-PI dataset. (Red: optimal, blue: second best).}
\label{table:lipi_global}
\resizebox{\columnwidth}{!}{
\begin{tabular}{ccccccc}
\toprule
\multirow{2}{*}{Methods} &
\multicolumn{6}{c}{Metrics} \\
\cmidrule(lr){2-7}
& $Q_{P}\uparrow$
& $Q_{S}\uparrow$
& $Q_{CB}\uparrow$
& $Q_{CV}\downarrow$
& $Q_{AB/F}\uparrow$
& MS-SSIM$\uparrow$ \\
\midrule
IRPol-Fuse & {\color[HTML]{FE0000}0.355} & {\color[HTML]{FE0000}0.805} & {\color[HTML]{FE0000}0.532} & {\color[HTML]{FE0000}171.583} & {\color[HTML]{FE0000}0.628} & {\color[HTML]{FE0000}0.930} \\
TIPFNet & 0.346 & {\color[HTML]{34CDF9}0.770} & {\color[HTML]{34CDF9}0.487} & 199.669 & {\color[HTML]{34CDF9}0.624} & {\color[HTML]{34CDF9}0.929} \\
CPIFuse & 0.176 & 0.475 & 0.218 & 216.936 & 0.248 & 0.744 \\
LFDT & 0.235 & 0.472 & 0.257 & 218.473 & 0.335 & 0.721 \\
LUT-Fuse & 0.149 & 0.367 & 0.240 & 211.443 & 0.266 & 0.624 \\
PIPFNet & 0.137 & 0.368 & 0.225 & 229.187 & 0.144 & 0.592 \\
DT-F & {\color[HTML]{34CDF9}0.354} & 0.501 & 0.332 & 322.379 & {\color[HTML]{34CDF9}0.624} & 0.928 \\
FusionMamba & 0.113 & 0.436 & 0.170 & 263.748 & 0.208 & 0.746 \\
CDDFuse & 0.295 & 0.555 & 0.262 & {\color[HTML]{34CDF9}186.820} & 0.354 & 0.825 \\
SeAFusion & 0.178 & 0.569 & 0.283 & 240.403 & 0.450 & 0.841 \\
\bottomrule
\end{tabular}
}
}
\end{table}

To further evaluate whether the main claims are supported by task-related measurements, Table~\ref{table:lipi_region} reports the region-aware metrics on LI-PI. IRPol-Fuse ranks first on both polarization texture metrics, achieving a Pol Grad Corr of 0.959 and a Pol Grad MAE of 0.012, which demonstrates its advantage in preserving polarization-derived texture gradients. It also obtains the second-best IR CNR and a competitive IR ROI Corr, suggesting that the proposed method maintains effective infrared target separability while recovering polarization structural details.

\begin{table}[!t]
\centering
\small
\setlength{\tabcolsep}{3pt}
{\rmfamily
\caption{Region-aware evaluation on the LI-PI dataset.}
\label{table:lipi_region}
\resizebox{\linewidth}{!}{
\begin{tabular}{ccccc}
\toprule
\multirow{2}{*}{Methods} &
\multicolumn{4}{c}{Metrics} \\
\cmidrule(lr){2-5}
& IR ROI Corr$\uparrow$
& IR CNR$\uparrow$
& Pol Grad Corr$\uparrow$
& Pol Grad MAE$\downarrow$ \\
\midrule
IRPol-Fuse & 0.886 & {\color[HTML]{34CDF9}3.306} & {\color[HTML]{FE0000}0.959} & {\color[HTML]{FE0000}0.012} \\
TIPFNet & 0.487 & 0.285 & 0.197 & 0.063 \\
CPIFuse & 0.848 & {\color[HTML]{FE0000}3.544} & 0.800 & 0.025 \\
LFDT & {\color[HTML]{FE0000}0.995} & 3.192 & 0.779 & 0.033 \\
LUT-Fuse & {\color[HTML]{34CDF9}0.964} & 3.065 & 0.788 & 0.034 \\
PIPFNet & 0.501 & 2.823 & 0.441 & 0.041 \\
DT-F & 0.534 & 2.207 & 0.859 & 0.028 \\
FusionMamba & 0.880 & 2.443 & {\color[HTML]{34CDF9}0.895} & 0.024 \\
CDDFuse & 0.941 & 2.257 & 0.894 & 0.029 \\
SeAFusion & 0.528 & 3.267 & 0.857 & {\color[HTML]{34CDF9}0.021} \\
\bottomrule
\end{tabular}
}
}
\end{table}

\begin{figure*}[!t]
    \centering
    \includegraphics[width=1\linewidth]{ 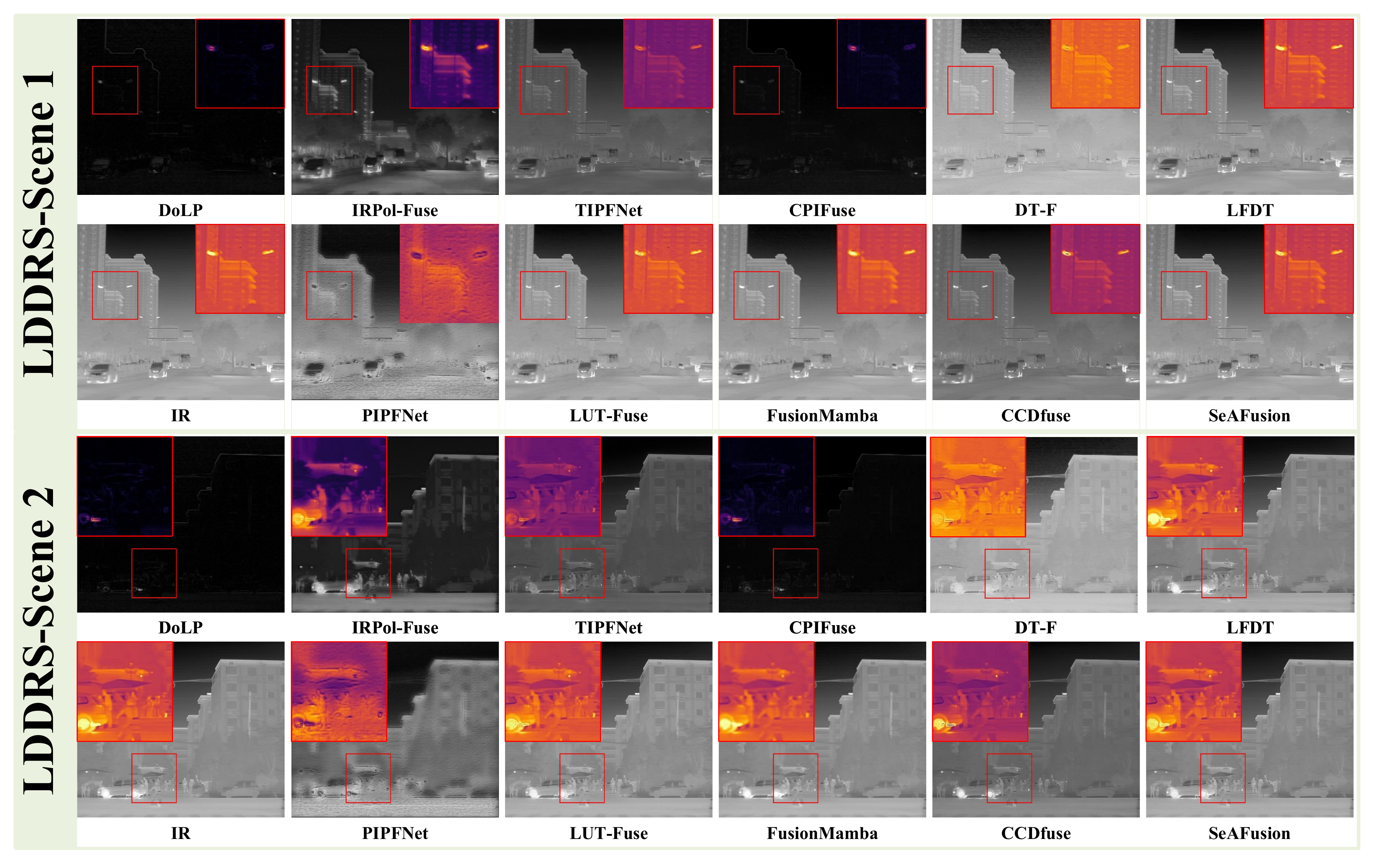}
    \vspace{-2em}
    \caption{Comparison of fusion results in two representative scenes from the LDDRS dataset.}
    \vspace{-2em}
    \label{fig:lddrs_results}
\end{figure*}

We further evaluate IRPol-Fuse on the external LDDRS dataset to examine generalization under conventional infrared--polarization imaging conditions. As shown in Fig.~\ref{fig:lddrs_results}, IRPol-Fuse preserves infrared saliency while retaining building edges and background structures. The quantitative results in Table~\ref{table:lddrs} show that IRPol-Fuse achieves the best $Q_{S}$ and $Q_{AB/F}$, and the second-best $Q_{P}$, $Q_{CB}$, and MS-SSIM. Although it does not obtain the lowest $Q_{CV}$, these results provide complementary evidence that the proposed energy--structure coordination strategy remains effective on external infrared--polarization data.

\begin{table}[!t]
\centering
\small
\setlength{\tabcolsep}{3pt}
{\rmfamily
\caption{Quantitative comparison results on the LDDRS dataset. (Red: optimal, blue: second best).}
\label{table:lddrs}
\resizebox{\columnwidth}{!}{
\begin{tabular}{ccccccc}
\toprule
\multirow{2}{*}{Methods} &
\multicolumn{6}{c}{Metrics} \\
\cmidrule(lr){2-7}
& $Q_{P}\uparrow$
& $Q_{S}\uparrow$
& $Q_{CB}\uparrow$
& $Q_{CV}\downarrow$
& $Q_{AB/F}\uparrow$
& MS-SSIM$\uparrow$ \\
\midrule
IRPol-Fuse &
{\color[HTML]{34CDF9}0.484} &
{\color[HTML]{FE0000}0.728} &
{\color[HTML]{34CDF9}0.499} &
393.908 &
{\color[HTML]{FE0000}0.534} &
{\color[HTML]{34CDF9}0.927} \\

TIPFNet &
0.329 &
0.579 &
0.320 &
982.441 &
{\color[HTML]{34CDF9}0.430} &
{\color[HTML]{FE0000}0.924} \\

CPIFuse &
{\color[HTML]{FE0000}0.509} &
0.676 &
{\color[HTML]{FE0000}0.507} &
1961.014 &
0.442 &
0.907 \\

LFDT &
0.374 &
0.468 &
0.224 &
79.728 &
0.408 &
0.809 \\

LUT-Fuse &
0.208 &
0.438 &
0.222 &
94.077 &
0.356 &
0.761 \\

PIPFNet &
0.094 &
0.311 &
0.335 &
1920.085 &
0.167 &
0.300 \\

DT-F &
0.343 &
0.418 &
0.256 &
346.184 &
0.397 &
{\color[HTML]{34CDF9}0.929} \\

FusionMamba &
0.123 &
0.485 &
0.250 &
{\color[HTML]{FE0000}75.322} &
0.227 &
0.880 \\

CDDFuse &
0.287 &
0.505 &
0.223 &
150.082 &
0.332 &
0.887 \\

SeAFusion &
0.268 &
{\color[HTML]{34CDF9}0.510} &
0.251 &
{\color[HTML]{34CDF9}77.722} &
0.374 &
0.895 \\
\bottomrule
\end{tabular}
}
}
\end{table}

\subsection{Ablation Analyses}
Ablation experiments are conducted on the LI-PI dataset to evaluate the contributions of the proposed modules. Starting from the complete IRPol-Fuse, different module combinations are removed to analyze the roles of cross-modal interaction, IHJ, and PTJ. The quantitative results are summarized in Table~\ref{table:ablation}.

\begin{table}[!t]
\caption{Ablation study results on the LI-PI dataset. (Red: optimal, blue: second best).}
\label{table:ablation}
\centering
\small
\setlength{\tabcolsep}{3pt}
\resizebox{\columnwidth}{!}{
\begin{tabular}{>{\centering\arraybackslash}m{2.5cm}cccccc}
\toprule
\multirow{2}{*}{Methods} &
\multicolumn{6}{c}{Metrics} \\
\cmidrule(lr){2-7}
& $Q_{P}\uparrow$
& $Q_{S}\uparrow$
& $Q_{CB}\uparrow$
& $Q_{CV}\downarrow$
& $Q_{AB/F}\uparrow$
& MS-SSIM$\uparrow$ \\
\midrule

IRPol-Fuse &
{\color[HTML]{FE0000}0.355} &
{\color[HTML]{FE0000}0.805} &
{\color[HTML]{FE0000}0.532} &
{\color[HTML]{FE0000}171.583} &
{\color[HTML]{FE0000}0.628} &
{\color[HTML]{FE0000}0.930} \\

W/O IHJ &
0.332 &
0.756 &
0.453 &
188.761 &
0.580 &
0.911 \\

W/O PTJ &
0.345 &
{\color[HTML]{34CDF9}0.804} &
{\color[HTML]{34CDF9}0.525} &
{\color[HTML]{34CDF9}179.074} &
{\color[HTML]{34CDF9}0.620} &
{\color[HTML]{34CDF9}0.928} \\

W/O Cross+IHJ &
{\color[HTML]{34CDF9}0.350} &
0.766 &
0.468 &
184.127 &
0.610 &
0.926 \\

W/O Cross+PTJ &
0.353 &
{\color[HTML]{34CDF9}0.805} &
0.523 &
189.215 &
0.616 &
0.926 \\

W/O IHJ+PTJ &
0.343 &
0.764 &
0.472 &
198.403 &
0.604 &
0.920 \\

Baseline &
0.346 &
0.762 &
0.457 &
188.551 &
0.591 &
0.917 \\

\bottomrule
\end{tabular}
}
\end{table}

Overall, the complete IRPol-Fuse achieves the best overall performance across all six global metrics. Compared with the baseline, the proposed model consistently improves $Q_{P}$, $Q_{S}$, $Q_{CB}$, $Q_{AB/F}$, MS-SSIM, and simultaneously obtains the lowest $Q_{CV}$, demonstrating that the proposed energy--structure coordinated design improves both structural consistency and perceptual fusion quality.

Removing IHJ leads to the most obvious degradation in almost all metrics, indicating that adaptive infrared reinjection is essential for preserving thermally salient targets during fusion. Removing PTJ also causes consistent performance degradation, especially in $Q_{CB}$, $Q_{AB/F}$, and MS-SSIM, suggesting that polarization texture restoration contributes substantially to structural fidelity and perceptual quality. Although several partial configurations, such as W/O Cross+PTJ, still maintain competitive structural consistency, none of them achieves balanced performance comparable to the complete model.

The results further demonstrate that the performance gain does not originate from a single module, but from the collaborative interaction among cross-modal feature learning, infrared highlight preservation, and polarization texture restoration. These observations are consistent with the proposed energy--structure coordinated fusion strategy.

\begin{figure*}[htbp]
    \centering
    \includegraphics[
        width=0.95\textwidth,
        trim=0 1.2cm 0 0,
        clip
    ]{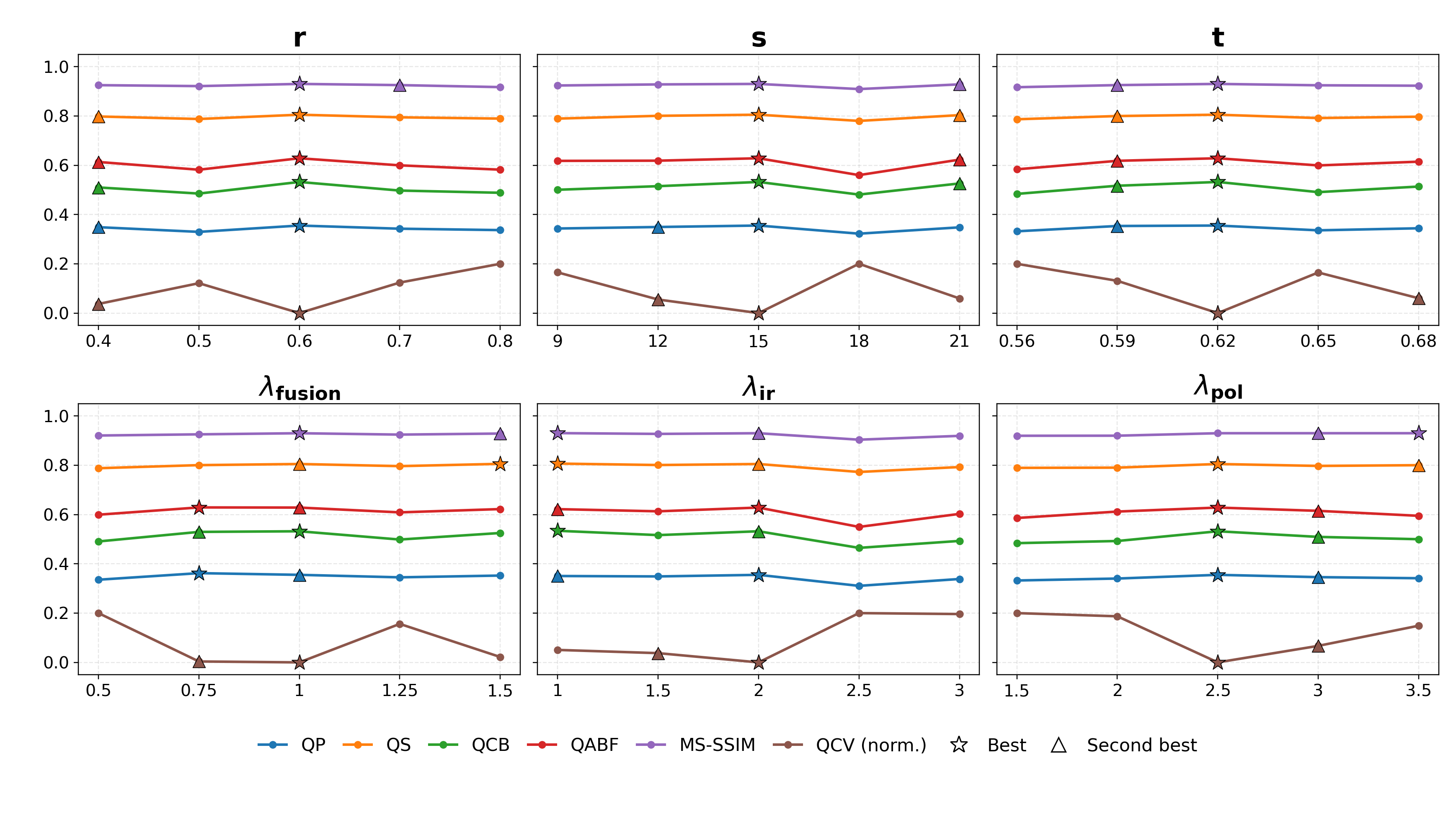}
    \vspace{-0.2cm}
    \caption{Hyperparameter sensitivity analysis on the LI-PI dataset. $Q_{CV}$ is min--max normalized for visualization.}
    \label{fig:sensitivity}
\end{figure*}

To further evaluate the robustness of the proposed framework, hyperparameter sensitivity analysis is performed for both the fixed IHJ hyperparameters ($r$, $s$, and $t$) and the three loss weights ($\lambda_{\mathrm{fusion}}$, $\lambda_{\mathrm{ir}}$, and $\lambda_{\mathrm{pol}}$). As shown in Fig.~\ref{fig:sensitivity}, IRPol-Fuse maintains relatively stable performance within a broad parameter range. The optimal configuration is obtained with $r=0.6$, $s=15$, $t=0.62$, $\lambda_{\mathrm{fusion}}=1.0$, $\lambda_{\mathrm{ir}}=2.0$, and $\lambda_{\mathrm{pol}}=2.5$. These results indicate that the proposed energy--structure coordination strategy is not overly sensitive to parameter selection.

\subsection{Generalization Experiment}

Although IRPol-Fuse is designed for infrared--polarization fusion, its energy--structure coordinated modeling is not restricted to a single data split. To provide complementary evidence for generalization, we conduct additional evaluations on two infrared--visible fusion datasets, M3FD and MSRS. IRPol-Fuse is directly transferred to infrared--visible fusion without modifying the network architecture or major hyperparameters.

\begin{figure*}[!t]
    \centering
    \centering
    \includegraphics[width=\linewidth]{ 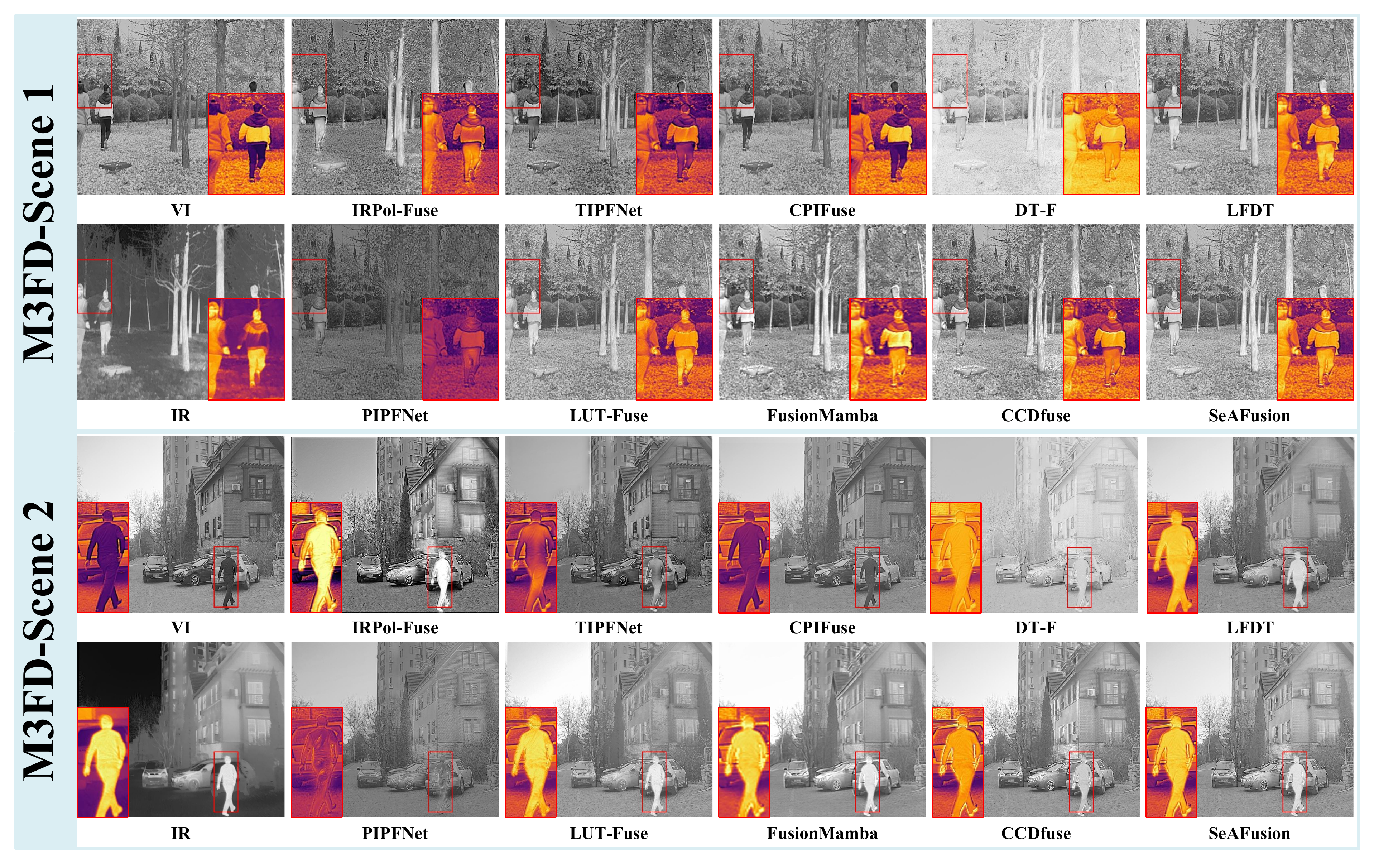}
    \caption{Generalization results on the M3FD dataset.}
    \label{fig:m3fd_generalization}
\end{figure*}

\begin{figure*}[!t]
    \centering
    \centering
    \includegraphics[width=\linewidth]{ 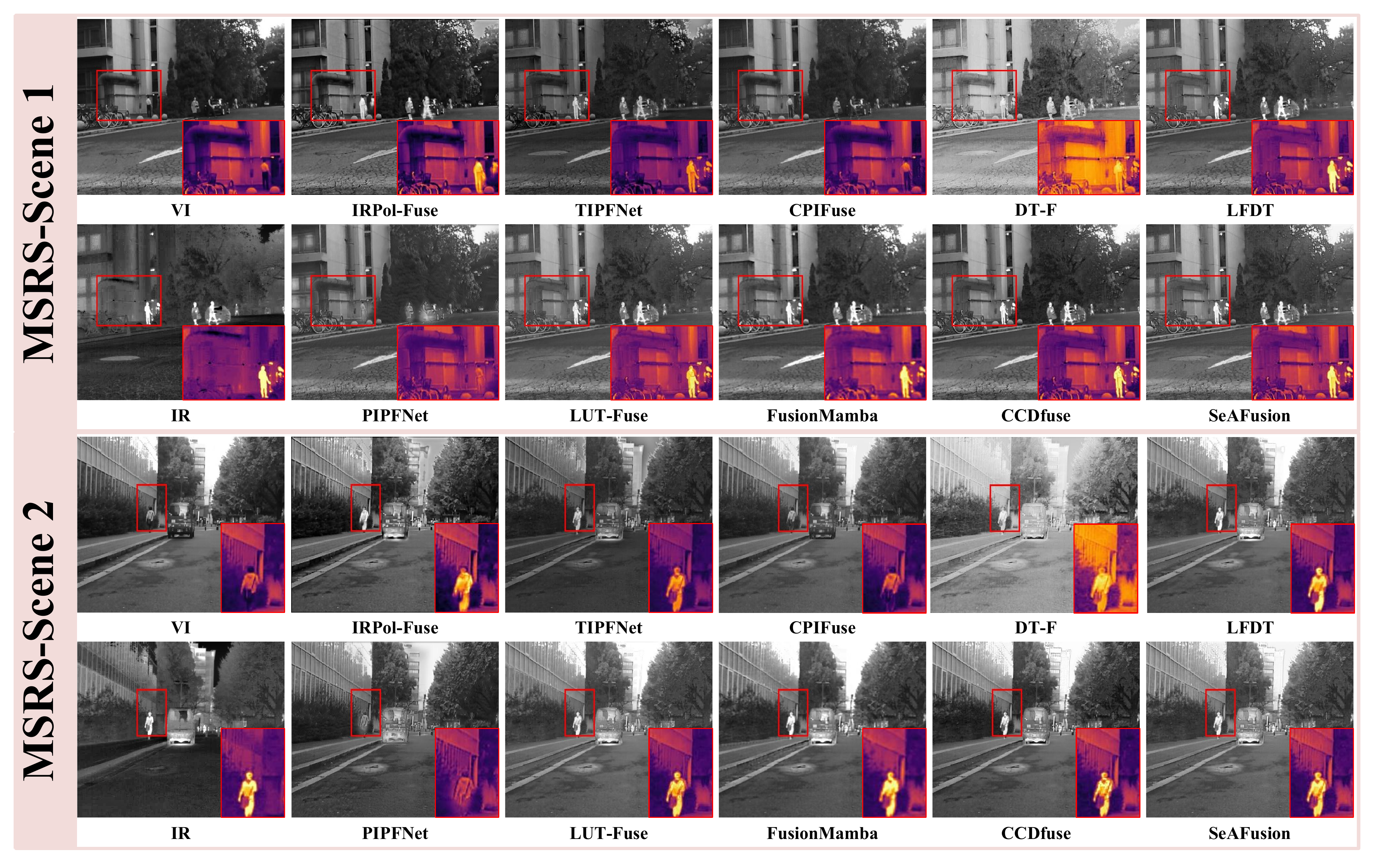}
    \caption{Generalization results on the MSRS dataset.}
    \label{fig:msrs_generalization}
\end{figure*}

As shown in Figs.~\ref{fig:m3fd_generalization} and~\ref{fig:msrs_generalization}, IRPol-Fuse preserves infrared target saliency while maintaining background structures and texture details on both datasets. The quantitative results in Tables~\ref{table:m3fd} and~\ref{table:msrs} further show that IRPol-Fuse achieves competitive performance on structure-sensitive metrics such as $Q_{S}$ and MS-SSIM. Although it does not outperform all infrared--visible fusion methods on every metric, these results provide complementary evidence that the proposed energy--structure coordination strategy has transferability beyond the original LI-PI split.

\begin{table}[!t]
\caption{Quantitative comparison results on the M3FD dataset.}
\label{table:m3fd}
\centering
\small
\setlength{\tabcolsep}{3pt}
\resizebox{\columnwidth}{!}{
\begin{tabular}{ccccccc}
\toprule
\multirow{2}{*}{Methods} &
\multicolumn{6}{c}{Metrics} \\
\cmidrule(lr){2-7}
& $Q_{P}\uparrow$
& $Q_{S}\uparrow$
& $Q_{CB}\uparrow$
& $Q_{CV}\downarrow$
& $Q_{AB/F}\uparrow$
& MS-SSIM$\uparrow$ \\
\midrule
IRPol-Fuse & 0.418 & {\color[HTML]{FE0000}0.889} & 0.522 & {\color[HTML]{34CDF9}422.508} & 0.523 & {\color[HTML]{FE0000}0.967} \\
TIPFNet & 0.445 & 0.818 & 0.524 & 713.855 & {\color[HTML]{FE0000}0.615} & 0.877 \\
CPIFuse & {\color[HTML]{34CDF9}0.475} & 0.853 & {\color[HTML]{FE0000}0.578} & 839.907 & 0.547 & 0.886 \\
LFDT & {\color[HTML]{FE0000}0.488} & 0.845 & 0.530 & 513.328 & {\color[HTML]{34CDF9}0.602} & 0.929 \\
LUT-Fuse & 0.423 & 0.820 & 0.512 & 550.227 & 0.563 & 0.913 \\
PIPFNet & 0.243 & 0.747 & 0.444 & 860.319 & 0.345 & 0.816 \\
DT-F & 0.337 & 0.701 & 0.356 & 533.248 & 0.455 & 0.908 \\
FusionMamba & 0.093 & 0.763 & 0.495 & 438.018 & 0.223 & 0.923 \\
CDDFuse & 0.474 & 0.861 & {\color[HTML]{34CDF9}0.537} & 427.048 & 0.594 & 0.952 \\
SeAFusion & 0.434 & {\color[HTML]{34CDF9}0.861} & 0.537 & {\color[HTML]{FE0000}419.169} & 0.600 & {\color[HTML]{34CDF9}0.959} \\
\bottomrule
\end{tabular}
}
\end{table}

\begin{table}[!t]
\caption{Quantitative comparison results on the MSRS dataset.}
\label{table:msrs}
\centering
\small
\setlength{\tabcolsep}{3pt}
\resizebox{\columnwidth}{!}{
\begin{tabular}{ccccccc}
\toprule
\multirow{2}{*}{Methods} &
\multicolumn{6}{c}{Metrics} \\
\cmidrule(lr){2-7}
& $Q_{P}\uparrow$
& $Q_{S}\uparrow$
& $Q_{CB}\uparrow$
& $Q_{CV}\downarrow$
& $Q_{AB/F}\uparrow$
& MS-SSIM$\uparrow$ \\
\midrule
IRPol-Fuse & 0.400 & {\color[HTML]{FE0000}0.885} & 0.527 & {\color[HTML]{FE0000}424.404} & 0.512 & {\color[HTML]{FE0000}0.961} \\
TIPFNet & 0.447 & 0.807 & 0.507 & 765.403 & {\color[HTML]{FE0000}0.626} & 0.861 \\
CPIFuse & {\color[HTML]{FE0000}0.480} & 0.852 & {\color[HTML]{FE0000}0.572} & 713.522 & 0.561 & 0.874 \\
LFDT & {\color[HTML]{34CDF9}0.476} & 0.842 & 0.521 & 537.411 & 0.600 & 0.916 \\
LUT-Fuse & 0.404 & 0.807 & 0.502 & 568.105 & 0.555 & 0.892 \\
PIPFNet & 0.238 & 0.744 & 0.441 & 801.091 & 0.347 & 0.804 \\
DT-F & 0.331 & 0.712 & 0.346 & 562.003 & 0.454 & 0.901 \\
FusionMamba & 0.101 & 0.751 & 0.486 & 505.569 & 0.231 & 0.911 \\
CDDFuse & 0.467 & {\color[HTML]{34CDF9}0.861} & {\color[HTML]{34CDF9}0.534} & {\color[HTML]{34CDF9}429.190} & 0.604 & 0.948 \\
SeAFusion & 0.426 & 0.859 & 0.533 & 448.187 & {\color[HTML]{34CDF9}0.605} & {\color[HTML]{34CDF9}0.952} \\
\bottomrule
\end{tabular}
}
\end{table}

\subsection{Downstream Object Detection Evaluation}

To further investigate whether the fused images preserve information useful for high-level visual perception, we conduct a downstream object detection experiment on the M3FD dataset. A fixed pretrained detector is applied directly to the fused images generated by different fusion methods without task-specific fine-tuning. During evaluation, the detector weights, input settings, confidence threshold, non-maximum suppression parameters, ground-truth annotations, and class definitions are kept identical for all competing methods. Therefore, differences in detection performance mainly reflect the influence of the fused images.

\begin{figure*}[!t]
    \centering
    \centering
    \includegraphics[width=\linewidth]{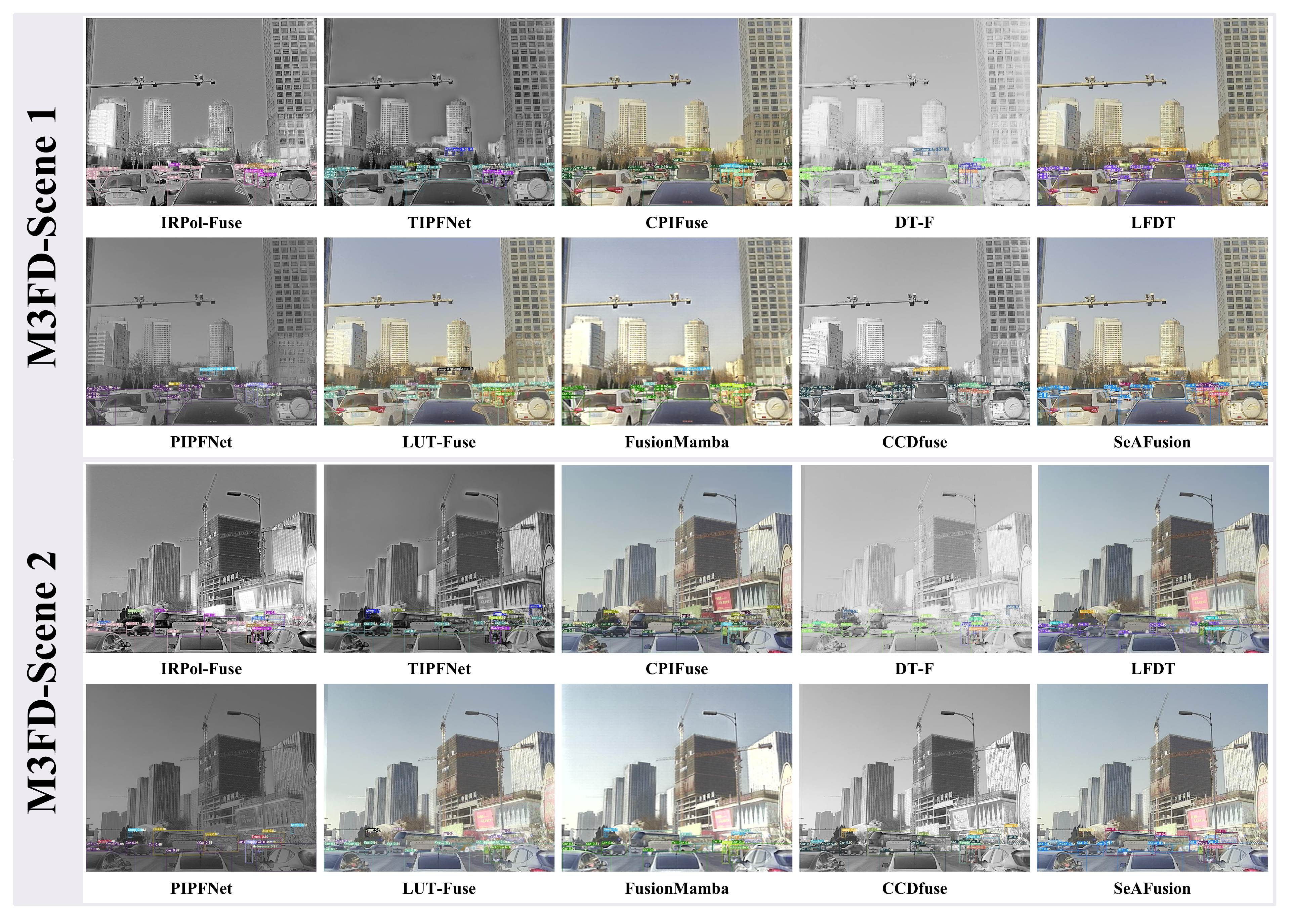}
    \caption{Downstream object detection evaluation on the M3FD dataset.}
    \label{fig:detection}
\end{figure*}

Precision, recall, and mAP@0.5 are adopted as evaluation metrics. As reported in Table~\ref{tab:detection}, IRPol-Fuse achieves a precision of 0.629, ranking second among all compared methods. Although its recall and mAP@0.5 are not the highest, the results demonstrate that IRPol-Fuse preserves useful target responses and structural cues for downstream detection without detector-specific optimization. Fig.~\ref{fig:detection} presents representative qualitative detection results on M3FD scenes. The same detector and visualization settings are applied to all fusion results, allowing the influence of different fused images on target recognition to be directly compared.

\begin{table}[htbp]
\centering
\small
\setlength{\tabcolsep}{6pt}
{\rmfamily
\caption{Downstream object detection results on the M3FD dataset.}
\label{tab:detection}
\resizebox{\columnwidth}{!}{
\begin{tabular}{>{\centering\arraybackslash}m{2.9cm}ccc}
\toprule
Methods & Precision$\uparrow$ & Recall$\uparrow$ & mAP@0.5$\uparrow$ \\
\midrule
IRPol-Fuse      & {\color[HTML]{34CDF9}0.629} & 0.484 & 0.563 \\
TIPFNet         & 0.615 & {\color[HTML]{34CDF9}0.547} & {\color[HTML]{34CDF9}0.605} \\
CPIFuse         & 0.597 & 0.515 & 0.560 \\
LFDT     & 0.625 & 0.528 & 0.575 \\
LUT-Fuse        & 0.580 & 0.544 & 0.576 \\
PIPFNet         & 0.519 & 0.537 & 0.537 \\
DT-F& {\color[HTML]{FE0000}0.638} & 0.479 & 0.542 \\
FusionMamba     & 0.565 & 0.496 & 0.518 \\
CDDFuse    & 0.599 & {\color[HTML]{FE0000}0.549} & 0.592 \\
SeAFusion       & 0.616 & 0.546 & {\color[HTML]{FE0000}0.610} \\
\bottomrule
\end{tabular}
}
}
\end{table}

\subsection{Computational Complexity Analysis}

To evaluate the computational cost of IRPol-Fuse, we compare the model complexity with representative fusion methods under the same input resolution. Table~\ref{tab:efficiency} reports the floating-point operations (FLOPs), the number of trainable parameters, and the average inference time.

\begin{table}[!t]
\centering
\small
\setlength{\tabcolsep}{3pt}
\caption{Comparison of computational complexity of different methods under the same input resolution.}
\label{tab:efficiency}
\resizebox{\columnwidth}{!}{
\begin{tabular}{>{\centering\arraybackslash}m{2.8cm}ccc}
\toprule
\multirow{2}{*}{Methods} &
\multicolumn{3}{c}{Metrics} \\
\cmidrule(lr){2-4}
& FLOPs (G)$\downarrow$
& Parameters (M)$\downarrow$
& Time (ms)$\downarrow$ \\
\midrule
IRPol-Fuse & 2608.007 & 4.553 & 515.279 \\
TIPFNet & 1569.841 & 4.151 & 768.695 \\
CPIFuse & 275.347 & 0.874 & 174.944 \\
LFDT & 3067.163 & 17.898 & 953.243 \\
LUT-Fuse & {\color[HTML]{FE0000}13.200} & {\color[HTML]{FE0000}0.008} & {\color[HTML]{FE0000}13.997} \\
PIPFNet & {\color[HTML]{34CDF9}121.133} & {\color[HTML]{34CDF9}0.034} & 455.377 \\
DT-F & 1173.224 & 0.834 & 1776.724 \\
FusionMamba & 688.729 & 85.489 & 306.040 \\
CDDFuse & 2983.823 & 1.186 & 1243.021 \\
SeAFusion & 277.797 & 0.167 & {\color[HTML]{34CDF9}74.086} \\
\bottomrule
\end{tabular}
}
\end{table}

As shown in Table~\ref{tab:efficiency}, IRPol-Fuse requires higher computational cost than several lightweight fusion methods because it incorporates cross-modal interaction together with the proposed IHJ and PTJ modules. Consequently, the additional feature extraction and adaptive refinement increase the overall computational burden.

Although IRPol-Fuse is not designed as a lightweight architecture, the complexity analysis provides a quantitative reference for its computational cost and facilitates a more comprehensive comparison with existing fusion methods.

\section{Conclusion}

This paper presented IRPol-Fuse, an infrared--polarization image fusion framework based on energy--structure coordination for challenging low-visibility scenarios. The proposed framework integrates PAF, IHJ, and  PTJ modules to jointly preserve thermally salient infrared targets and polarization-derived structural details. By combining adaptive infrared reinjection with mask-guided polarization texture restoration, IRPol-Fuse achieves coordinated enhancement of energy information and fine structural features.

Extensive experiments on the LI-PI dataset demonstrate that IRPol-Fuse consistently outperforms representative comparison methods in both qualitative evaluation and quantitative analysis. Additional region-aware metrics further verify its effectiveness in infrared target preservation and polarization texture recovery. Cross-dataset evaluations on LDDRS, MSRS, and M3FD, together with downstream object detection experiments on M3FD, provide complementary evidence for the transferability and practical applicability of the proposed energy--structure coordination strategy.

Nevertheless, the current study is still limited by the scale of the available infrared--polarization dataset. In future work, we plan to construct a larger and more diverse infrared--polarization benchmark and further investigate more efficient architectures to improve computational efficiency while maintaining high-quality fusion performance in real-world applications.

\section{Acknowledgements}
This research was supported by the Natural Science Foundation of Guangdong Province (No. 2024A1515011880), the Basic and Applied Basic Research of Guangdong Province (No. 2023A1515140077), the Research Fund of Guangdong-HongKong-Macao Joint Laboratory for Intelligent Micro-Nano Optoelectronic Technology (No. 2020B1212030010).

\printcredits

\bibliographystyle{cas-model2-names}

\bibliography{cas-refs}

\begin{thebibliography}{36}
\expandafter\ifx\csname natexlab\endcsname\relax\def\natexlab#1{#1}\fi
\providecommand{\url}[1]{\texttt{#1}}
\providecommand{\href}[2]{#2}
\providecommand{\path}[1]{#1}
\providecommand{\DOIprefix}{doi:}
\providecommand{\ArXivprefix}{arXiv:}
\providecommand{\URLprefix}{URL: }
\providecommand{\Pubmedprefix}{pmid:}
\providecommand{\doi}[1]{\href{http://dx.doi.org/#1}{\path{#1}}}
\providecommand{\Pubmed}[1]{\href{pmid:#1}{\path{#1}}}
\providecommand{\bibinfo}[2]{#2}
\ifx\xfnm\relax \def\xfnm[#1]{\unskip,\space#1}\fi
\bibitem[{Tyo et~al.(2007)Tyo, Ratliff, Boger, Black, Bowers and Fetrow}]{54}
\bibinfo{author}{Tyo, J.S.}, \bibinfo{author}{Ratliff, B.M.}, \bibinfo{author}{Boger, J.K.}, \bibinfo{author}{Black, W.T.}, \bibinfo{author}{Bowers, D.L.}, \bibinfo{author}{Fetrow, M.P.}, \bibinfo{year}{2007}.
\newblock \bibinfo{title}{The effects of thermal equilibrium and contrast in {LWIR} polarimetric images}.
\newblock \bibinfo{journal}{Optics Express} \bibinfo{volume}{15}, \bibinfo{pages}{15161--15167}.
\newblock \DOIprefix\doi{10.1364/OE.15.015161}.
\bibitem[{Wang et~al.(2023)Wang, Jiao, Hao, Shan, Song and Huang}]{1}
\bibinfo{author}{Wang, H.F.}, \bibinfo{author}{Jiao, Y.M.}, \bibinfo{author}{Hao, T.}, \bibinfo{author}{Shan, Y.H.}, \bibinfo{author}{Song, S.Z.}, \bibinfo{author}{Huang, H.}, \bibinfo{year}{2023}.
\newblock \bibinfo{title}{Low-visibility vehicle-road environment perception based on the multi-modal visual features fusion of polarization and infrared}.
\newblock \bibinfo{journal}{IEEE Transactions on Intelligent Transportation Systems} \bibinfo{volume}{24}, \bibinfo{pages}{11997--12013}.
\newblock \DOIprefix\doi{10.1109/TITS.2023.3286541}.
\bibitem[{Wu et~al.(2024)Wu, Zhou, Wang, Peng, Lin, Cao and Huang}]{2}
\bibinfo{author}{Wu, X.}, \bibinfo{author}{Zhou, B.}, \bibinfo{author}{Wang, X.}, \bibinfo{author}{Peng, J.}, \bibinfo{author}{Lin, P.}, \bibinfo{author}{Cao, R.}, \bibinfo{author}{Huang, F.}, \bibinfo{year}{2024}.
\newblock \bibinfo{title}{{SwinIPISR}: A super-resolution method for infrared polarization imaging sensors via swin transformer}.
\newblock \bibinfo{journal}{IEEE Sensors Journal} \bibinfo{volume}{24}, \bibinfo{pages}{468--477}.
\newblock \DOIprefix\doi{10.1109/JSEN.2023.3331578}.
\bibitem[{Meng et~al.(2025)Meng, Fan, Hu, Zhang and Zhao}]{3}
\bibinfo{author}{Meng, X.}, \bibinfo{author}{Fan, Z.}, \bibinfo{author}{Hu, Y.}, \bibinfo{author}{Zhang, Y.}, \bibinfo{author}{Zhao, C.}, \bibinfo{year}{2025}.
\newblock \bibinfo{title}{Infrared polarization reconstruction method for ship target detection with latent low-rank representation and salient region extraction}.
\newblock \bibinfo{journal}{IEEE Sensors Journal} \bibinfo{volume}{25}, \bibinfo{pages}{22090--22099}.
\newblock \DOIprefix\doi{10.1109/JSEN.2025.3564007}.
\bibitem[{Wang et~al.(2024)Wang, Ma, Huang, Fan, Li and Li}]{45}
\bibinfo{author}{Wang, G.}, \bibinfo{author}{Ma, Y.}, \bibinfo{author}{Huang, J.}, \bibinfo{author}{Fan, F.}, \bibinfo{author}{Li, H.}, \bibinfo{author}{Li, Z.}, \bibinfo{year}{2024}.
\newblock \bibinfo{title}{Instance segmentation of pigs in infrared images based on {INPC} model}.
\newblock \bibinfo{journal}{Infrared Physics \& Technology} \bibinfo{volume}{141}, \bibinfo{pages}{105491}.
\newblock \DOIprefix\doi{10.1016/j.infrared.2024.105491}.
\bibitem[{Wang et~al.(2025)Wang, Ma, Huang, Fan and Wang}]{46}
\bibinfo{author}{Wang, G.}, \bibinfo{author}{Ma, Y.}, \bibinfo{author}{Huang, J.}, \bibinfo{author}{Fan, F.}, \bibinfo{author}{Wang, Z.}, \bibinfo{year}{2025}.
\newblock \bibinfo{title}{Measurement of pig body temperature based on ear segmentation and multifactor infrared temperature compensation}.
\newblock \bibinfo{journal}{IEEE Transactions on Instrumentation and Measurement} \bibinfo{volume}{74}, \bibinfo{pages}{1--15}.
\newblock \DOIprefix\doi{10.1109/TIM.2025.3538062}.
\bibitem[{Zhu et~al.(2025)Zhu, Ma, Fan, Huang and Wang}]{47}
\bibinfo{author}{Zhu, Y.}, \bibinfo{author}{Ma, Y.}, \bibinfo{author}{Fan, F.}, \bibinfo{author}{Huang, J.}, \bibinfo{author}{Wang, G.}, \bibinfo{year}{2025}.
\newblock \bibinfo{title}{Shifting neighbors within temporal contexts for slow-moving infrared small target detection}.
\newblock \bibinfo{journal}{IEEE Signal Processing Letters} \bibinfo{volume}{32}, \bibinfo{pages}{3999--4003}.
\newblock \DOIprefix\doi{10.1109/LSP.2025.3616637}.
\bibitem[{Wang et~al.(2026)Wang, Ma, Huang, Du, Fan, Zhao and Lu}]{48}
\bibinfo{author}{Wang, G.}, \bibinfo{author}{Ma, Y.}, \bibinfo{author}{Huang, J.}, \bibinfo{author}{Du, Y.}, \bibinfo{author}{Fan, F.}, \bibinfo{author}{Zhao, Z.}, \bibinfo{author}{Lu, Y.}, \bibinfo{year}{2026}.
\newblock \bibinfo{title}{Unleashing the potential of {Mamba}: A novel approach for low-light image enhancement}.
\newblock \bibinfo{journal}{Knowledge-Based Systems} \bibinfo{volume}{346}, \bibinfo{pages}{116205}.
\newblock \DOIprefix\doi{10.1016/j.knosys.2026.116205}.
\bibitem[{Lu et~al.(2026)Lu, Huang, Ma, Fan, Wu and Wang}]{49}
\bibinfo{author}{Lu, Y.}, \bibinfo{author}{Huang, J.}, \bibinfo{author}{Ma, Y.}, \bibinfo{author}{Fan, F.}, \bibinfo{author}{Wu, K.}, \bibinfo{author}{Wang, G.}, \bibinfo{year}{2026}.
\newblock \bibinfo{title}{Frequency-aware retinex-wavelet decomposition hybrid network and luminance-guided transformers for low-light image enhancement}.
\newblock \bibinfo{journal}{Optics \& Laser Technology} \bibinfo{volume}{194}, \bibinfo{pages}{114432}.
\newblock \DOIprefix\doi{10.1016/j.optlastec.2025.114432}.
\bibitem[{Wang et~al.(2026)Wang, Ma, Huang, Wu, Tang, Cai and Fan}]{50}
\bibinfo{author}{Wang, G.}, \bibinfo{author}{Ma, Y.}, \bibinfo{author}{Huang, J.}, \bibinfo{author}{Wu, K.}, \bibinfo{author}{Tang, L.}, \bibinfo{author}{Cai, Z.}, \bibinfo{author}{Fan, F.}, \bibinfo{year}{2026}.
\newblock \bibinfo{title}{{CLTE}: Infrared and visible image fusion through cross-domain learning and texture-contrast enhancement}.
\newblock \bibinfo{journal}{Infrared Physics \& Technology} \bibinfo{volume}{152}, \bibinfo{pages}{106230}.
\newblock \DOIprefix\doi{10.1016/j.infrared.2025.106230}.
\bibitem[{Zhao et~al.(2026)Zhao, Ma, Huang, Wu, Wang and Fan}]{51}
\bibinfo{author}{Zhao, Z.}, \bibinfo{author}{Ma, Y.}, \bibinfo{author}{Huang, J.}, \bibinfo{author}{Wu, K.}, \bibinfo{author}{Wang, G.}, \bibinfo{author}{Fan, F.}, \bibinfo{year}{2026}.
\newblock \bibinfo{title}{{CrossMamba}: Cross-modal features mixing via {Mamba} for infrared and visible image fusion}.
\newblock \bibinfo{journal}{Infrared Physics \& Technology} \bibinfo{volume}{152}, \bibinfo{pages}{106234}.
\newblock \DOIprefix\doi{10.1016/j.infrared.2025.106234}.
\bibitem[{Cai et~al.(2025)Cai, Ma, Huang, Cai, Wang and Fan}]{52}
\bibinfo{author}{Cai, Z.}, \bibinfo{author}{Ma, Y.}, \bibinfo{author}{Huang, J.}, \bibinfo{author}{Cai, Z.}, \bibinfo{author}{Wang, G.}, \bibinfo{author}{Fan, F.}, \bibinfo{year}{2025}.
\newblock \bibinfo{title}{{WaveFusion}: Wave-aware feature mixing network for multi-sensor fusion of infrared and visible images}.
\newblock \bibinfo{journal}{IEEE Sensors Journal} \bibinfo{volume}{25}, \bibinfo{pages}{25146--25159}.
\newblock \DOIprefix\doi{10.1109/JSEN.2025.3571945}.
\bibitem[{Duan et~al.(2024)Duan, Liu, Hao, Chen, Zheng and Jia}]{4}
\bibinfo{author}{Duan, J.}, \bibinfo{author}{Liu, J.}, \bibinfo{author}{Hao, Y.}, \bibinfo{author}{Chen, G.}, \bibinfo{author}{Zheng, Y.}, \bibinfo{author}{Jia, L.}, \bibinfo{year}{2024}.
\newblock \bibinfo{title}{Joint target geometry and polarization properties for polarization image fusion}.
\newblock \bibinfo{journal}{Optics and Lasers in Engineering} \bibinfo{volume}{178}, \bibinfo{pages}{108176}.
\newblock \DOIprefix\doi{10.1016/j.optlaseng.2024.108176}.
\bibitem[{Chen and Wolff(1998)}]{5}
\bibinfo{author}{Chen, H.}, \bibinfo{author}{Wolff, L.B.}, \bibinfo{year}{1998}.
\newblock \bibinfo{title}{Polarization phase-based method for material classification in computer vision}.
\newblock \bibinfo{journal}{International Journal of Computer Vision} \bibinfo{volume}{28}, \bibinfo{pages}{73--83}.
\newblock \DOIprefix\doi{10.1023/A:1008054731537}.
\bibitem[{Yan et~al.(2026)Yan, Liu, Li, Zhang, Zhang and Zhang}]{6}
\bibinfo{author}{Yan, X.}, \bibinfo{author}{Liu, K.}, \bibinfo{author}{Li, J.}, \bibinfo{author}{Zhang, Y.}, \bibinfo{author}{Zhang, Y.}, \bibinfo{author}{Zhang, C.}, \bibinfo{year}{2026}.
\newblock \bibinfo{title}{Drone detection network based on {RGB}-thermal imaging multimodal fusion}.
\newblock \bibinfo{journal}{Infrared Physics \& Technology} \bibinfo{volume}{155}, \bibinfo{pages}{106426}.
\newblock \DOIprefix\doi{10.1016/j.infrared.2026.106426}.
\bibitem[{Zhao et~al.(2026)Zhao, Zhang, Li, Yue and Shi}]{8}
\bibinfo{author}{Zhao, S.}, \bibinfo{author}{Zhang, C.}, \bibinfo{author}{Li, H.}, \bibinfo{author}{Yue, M.}, \bibinfo{author}{Shi, R.}, \bibinfo{year}{2026}.
\newblock \bibinfo{title}{{EMFusionNet}: Attention-guided wavelet {Mamba} for cross-modal fusion in nighttime infrared--visible images}.
\newblock \bibinfo{journal}{Infrared Physics \& Technology} \bibinfo{volume}{153}, \bibinfo{pages}{106323}.
\newblock \DOIprefix\doi{10.1016/j.infrared.2025.106323}.
\bibitem[{Jie et~al.(2025)Jie, Xu, Li, Zhou, Lv and Li}]{9}
\bibinfo{author}{Jie, Y.}, \bibinfo{author}{Xu, Y.}, \bibinfo{author}{Li, X.}, \bibinfo{author}{Zhou, F.}, \bibinfo{author}{Lv, J.}, \bibinfo{author}{Li, H.}, \bibinfo{year}{2025}.
\newblock \bibinfo{title}{{FS-Diff}: Semantic guidance and clarity-aware simultaneous multimodal image fusion and super-resolution}.
\newblock \bibinfo{journal}{Information Fusion} \bibinfo{volume}{121}, \bibinfo{pages}{103146}.
\newblock \DOIprefix\doi{10.1016/j.inffus.2025.103146}.
\bibitem[{Li et~al.(2022)Li, Qi, Zhuang, Yang and Gao}]{7}
\bibinfo{author}{Li, K.}, \bibinfo{author}{Qi, M.}, \bibinfo{author}{Zhuang, S.}, \bibinfo{author}{Yang, Y.}, \bibinfo{author}{Gao, J.}, \bibinfo{year}{2022}.
\newblock \bibinfo{title}{{TIPFNet}: A transformer-based infrared polarization image fusion network}.
\newblock \bibinfo{journal}{Optics Letters} \bibinfo{volume}{47}, \bibinfo{pages}{4255--4258}.
\newblock \DOIprefix\doi{10.1364/OL.466191}.
\bibitem[{Luo et~al.(2023)Luo, Fu, Yang, Cao and Cao}]{10}
\bibinfo{author}{Luo, X.}, \bibinfo{author}{Fu, G.}, \bibinfo{author}{Yang, J.}, \bibinfo{author}{Cao, Y.}, \bibinfo{author}{Cao, Y.}, \bibinfo{year}{2023}.
\newblock \bibinfo{title}{Multi-modal image fusion via deep laplacian pyramid hybrid network}.
\newblock \bibinfo{journal}{IEEE Transactions on Circuits and Systems for Video Technology} \bibinfo{volume}{33}, \bibinfo{pages}{7354--7369}.
\newblock \DOIprefix\doi{10.1109/TCSVT.2023.3281462}.
\bibitem[{Li et~al.(2023)Li, Tan, Zhou, Wang and Li}]{11}
\bibinfo{author}{Li, X.}, \bibinfo{author}{Tan, H.}, \bibinfo{author}{Zhou, F.}, \bibinfo{author}{Wang, G.}, \bibinfo{author}{Li, X.}, \bibinfo{year}{2023}.
\newblock \bibinfo{title}{Infrared and visible image fusion based on domain transform filtering and sparse representation}.
\newblock \bibinfo{journal}{Infrared Physics \& Technology} \bibinfo{volume}{131}, \bibinfo{pages}{104701}.
\newblock \DOIprefix\doi{10.1016/j.infrared.2023.104701}.
\bibitem[{Liu et~al.(2017)Liu, Qi and Ding}]{13}
\bibinfo{author}{Liu, C.H.}, \bibinfo{author}{Qi, Y.}, \bibinfo{author}{Ding, W.R.}, \bibinfo{year}{2017}.
\newblock \bibinfo{title}{Infrared and visible image fusion method based on saliency detection in sparse domain}.
\newblock \bibinfo{journal}{Infrared Physics \& Technology} \bibinfo{volume}{83}, \bibinfo{pages}{94--102}.
\newblock \DOIprefix\doi{10.1016/j.infrared.2017.04.018}.
\bibitem[{Li et~al.(2024)Li, Qi, Zhuang and Liu}]{12}
\bibinfo{author}{Li, K.}, \bibinfo{author}{Qi, M.}, \bibinfo{author}{Zhuang, S.}, \bibinfo{author}{Liu, Y.}, \bibinfo{year}{2024}.
\newblock \bibinfo{title}{Polarized prior guided fusion network for infrared polarization images}.
\newblock \bibinfo{journal}{IEEE Transactions on Geoscience and Remote Sensing} \bibinfo{volume}{62}, \bibinfo{pages}{1--17}.
\newblock \DOIprefix\doi{10.1109/TGRS.2024.3389976}.
\bibitem[{Ting et~al.(2021)Ting, Wu, Hu and Zhang}]{14}
\bibinfo{author}{Ting, J.}, \bibinfo{author}{Wu, X.}, \bibinfo{author}{Hu, K.}, \bibinfo{author}{Zhang, H.}, \bibinfo{year}{2021}.
\newblock \bibinfo{title}{Deep snapshot {HDR} reconstruction based on the polarization camera}, in: \bibinfo{booktitle}{2021 IEEE International Conference on Image Processing (ICIP)}, pp. \bibinfo{pages}{1769--1773}.
\newblock \DOIprefix\doi{10.1109/ICIP42928.2021.9506314}.
\bibitem[{Ma et~al.(2019)Ma, Yu, Liang, Li and Jiang}]{18}
\bibinfo{author}{Ma, J.}, \bibinfo{author}{Yu, W.}, \bibinfo{author}{Liang, P.}, \bibinfo{author}{Li, C.}, \bibinfo{author}{Jiang, J.}, \bibinfo{year}{2019}.
\newblock \bibinfo{title}{{FusionGAN}: A generative adversarial network for infrared and visible image fusion}.
\newblock \bibinfo{journal}{Information Fusion} \bibinfo{volume}{48}, \bibinfo{pages}{11--26}.
\newblock \DOIprefix\doi{10.1016/j.inffus.2018.09.004}.
\bibitem[{Zhou et~al.(2024)Zhou, Zeng, Lin, Li, Ali~Shah, Liu, Guo and Guo}]{21}
\bibinfo{author}{Zhou, H.}, \bibinfo{author}{Zeng, X.}, \bibinfo{author}{Lin, B.}, \bibinfo{author}{Li, D.}, \bibinfo{author}{Ali~Shah, S.A.}, \bibinfo{author}{Liu, B.}, \bibinfo{author}{Guo, K.}, \bibinfo{author}{Guo, Z.}, \bibinfo{year}{2024}.
\newblock \bibinfo{title}{Polarization motivating high-performance weak targets' imaging based on a dual-discriminator {GAN}}.
\newblock \bibinfo{journal}{Optics Express} \bibinfo{volume}{32}, \bibinfo{pages}{3835--3851}.
\newblock \DOIprefix\doi{10.1364/OE.504918}.
\bibitem[{Cui et~al.(2023)Cui, Chen, Gu, Yang and Shi}]{22}
\bibinfo{author}{Cui, S.}, \bibinfo{author}{Chen, W.}, \bibinfo{author}{Gu, W.}, \bibinfo{author}{Yang, L.}, \bibinfo{author}{Shi, X.}, \bibinfo{year}{2023}.
\newblock \bibinfo{title}{{SiamC Transformer}: Siamese coupling swin transformer multi-scale semantic segmentation network for vegetation extraction under shadow conditions}.
\newblock \bibinfo{journal}{Computers and Electronics in Agriculture} \bibinfo{volume}{213}, \bibinfo{pages}{108245}.
\newblock \DOIprefix\doi{10.1016/j.compag.2023.108245}.
\bibitem[{Liu et~al.(2024)Liu, Li, Dian and Song}]{23}
\bibinfo{author}{Liu, J.}, \bibinfo{author}{Li, S.}, \bibinfo{author}{Dian, R.}, \bibinfo{author}{Song, Z.}, \bibinfo{year}{2024}.
\newblock \bibinfo{title}{{DT-F Transformer}: Dual transpose fusion transformer for polarization image fusion}.
\newblock \bibinfo{journal}{Information Fusion} \bibinfo{volume}{106}, \bibinfo{pages}{102274}.
\newblock \DOIprefix\doi{10.1016/j.inffus.2024.102274}.
\bibitem[{Schechner et~al.(2003)Schechner, Narasimhan and Nayar}]{26}
\bibinfo{author}{Schechner, Y.Y.}, \bibinfo{author}{Narasimhan, S.G.}, \bibinfo{author}{Nayar, S.K.}, \bibinfo{year}{2003}.
\newblock \bibinfo{title}{Polarization-based vision through haze}.
\newblock \bibinfo{journal}{Applied Optics} \bibinfo{volume}{42}, \bibinfo{pages}{511--525}.
\newblock \DOIprefix\doi{10.1364/AO.42.000511}.
\bibitem[{Liu et~al.(2025)Liu, Shan, Ding, Qian, Yang and Cheng}]{28}
\bibinfo{author}{Liu, S.}, \bibinfo{author}{Shan, J.}, \bibinfo{author}{Ding, R.}, \bibinfo{author}{Qian, J.}, \bibinfo{author}{Yang, Z.}, \bibinfo{author}{Cheng, X.}, \bibinfo{year}{2025}.
\newblock \bibinfo{title}{An algorithm for infrared polarized light imaging band preference under matrix fourier optics optimization}, in: \bibinfo{booktitle}{2025 2nd International Conference on Digital Image Processing and Computer Applications (DIPCA)}, pp. \bibinfo{pages}{162--165}.
\newblock \DOIprefix\doi{10.1109/DIPCA65051.2025.11042308}.
\bibitem[{Li et~al.(2020)Li, Zhao, Pan, Kong and Chan}]{33}
\bibinfo{author}{Li, N.}, \bibinfo{author}{Zhao, Y.}, \bibinfo{author}{Pan, Q.}, \bibinfo{author}{Kong, S.G.}, \bibinfo{author}{Chan, J.C.W.}, \bibinfo{year}{2020}.
\newblock \bibinfo{title}{Full-time monocular road detection using zero-distribution prior of angle of polarization}, in: \bibinfo{booktitle}{Computer Vision -- ECCV 2020}, pp. \bibinfo{pages}{457--473}.
\newblock \DOIprefix\doi{10.1007/978-3-030-58595-2_28}.
\bibitem[{Luo et~al.(2025)Luo, Zhang and Li}]{34}
\bibinfo{author}{Luo, Y.}, \bibinfo{author}{Zhang, J.}, \bibinfo{author}{Li, C.}, \bibinfo{year}{2025}.
\newblock \bibinfo{title}{{CPIFuse}: Toward realistic color and enhanced textures in color polarization image fusion}.
\newblock \bibinfo{journal}{Information Fusion} \bibinfo{volume}{120}, \bibinfo{pages}{103111}.
\newblock \DOIprefix\doi{10.1016/j.inffus.2025.103111}.
\bibitem[{Yang et~al.(2025)Yang, Jiang, Pan, Yu, Gui and Gui}]{37}
\bibinfo{author}{Yang, B.}, \bibinfo{author}{Jiang, Z.}, \bibinfo{author}{Pan, D.}, \bibinfo{author}{Yu, H.}, \bibinfo{author}{Gui, G.}, \bibinfo{author}{Gui, W.}, \bibinfo{year}{2025}.
\newblock \bibinfo{title}{{LFDT-Fusion}: A latent feature-guided diffusion transformer model for general image fusion}.
\newblock \bibinfo{journal}{Information Fusion} \bibinfo{volume}{113}, \bibinfo{pages}{102639}.
\newblock \DOIprefix\doi{10.1016/j.inffus.2024.102639}.
\bibitem[{Xie et~al.(2024)Xie, Cui, Tan, Zheng and Yu}]{38}
\bibinfo{author}{Xie, X.}, \bibinfo{author}{Cui, Y.}, \bibinfo{author}{Tan, T.}, \bibinfo{author}{Zheng, X.}, \bibinfo{author}{Yu, Z.}, \bibinfo{year}{2024}.
\newblock \bibinfo{title}{{FusionMamba}: Dynamic feature enhancement for multimodal image fusion with {Mamba}}.
\newblock \bibinfo{journal}{Visual Intelligence} \bibinfo{volume}{2}, \bibinfo{pages}{37}.
\newblock \DOIprefix\doi{10.1007/s44267-024-00072-9}.
\bibitem[{Zhao et~al.(2023)Zhao, Bai, Zhang, Zhang, Xu, Lin, Timofte and Van~Gool}]{39}
\bibinfo{author}{Zhao, Z.}, \bibinfo{author}{Bai, H.}, \bibinfo{author}{Zhang, J.}, \bibinfo{author}{Zhang, Y.}, \bibinfo{author}{Xu, S.}, \bibinfo{author}{Lin, Z.}, \bibinfo{author}{Timofte, R.}, \bibinfo{author}{Van~Gool, L.}, \bibinfo{year}{2023}.
\newblock \bibinfo{title}{{CDDFuse}: Correlation-driven dual-branch feature decomposition for multi-modality image fusion}, in: \bibinfo{booktitle}{Proceedings of the IEEE/CVF Conference on Computer Vision and Pattern Recognition}, pp. \bibinfo{pages}{5906--5916}.
\newblock \DOIprefix\doi{10.1109/CVPR52729.2023.00572}.
\bibitem[{Yi et~al.(2025)Yi, Zhang, Xiang, Yan, Xu and Ma}]{40}
\bibinfo{author}{Yi, X.}, \bibinfo{author}{Zhang, Y.}, \bibinfo{author}{Xiang, X.}, \bibinfo{author}{Yan, Q.}, \bibinfo{author}{Xu, H.}, \bibinfo{author}{Ma, J.}, \bibinfo{year}{2025}.
\newblock \bibinfo{title}{{LUT-Fuse}: Towards extremely fast infrared and visible image fusion via distillation to learnable look-up tables}, in: \bibinfo{booktitle}{Proceedings of the IEEE/CVF International Conference on Computer Vision}, pp. \bibinfo{pages}{14559--14568}.
\newblock \DOIprefix\doi{10.1109/ICCV51701.2025.01351}.
\bibitem[{Tang et~al.(2022)Tang, Yuan and Ma}]{41}
\bibinfo{author}{Tang, L.}, \bibinfo{author}{Yuan, J.}, \bibinfo{author}{Ma, J.}, \bibinfo{year}{2022}.
\newblock \bibinfo{title}{Image fusion in the loop of high-level vision tasks: A semantic-aware real-time infrared and visible image fusion network}.
\newblock \bibinfo{journal}{Information Fusion} \bibinfo{volume}{82}, \bibinfo{pages}{28--42}.
\newblock \DOIprefix\doi{10.1016/j.inffus.2021.12.004}.

\end{thebibliography}


\end{document}